\documentclass[10pt,twocolumn,letterpaper]{article}

\usepackage{cvpr}              
\usepackage{colortbl} 
\usepackage{multirow} 
\usepackage{xcolor}   
\usepackage{array}    
\usepackage{graphicx}
\usepackage{caption}
\usepackage{algorithm}
\usepackage{algpseudocode}
\usepackage{fancyhdr}
\definecolor{lightpurple}{rgb}{0.85, 0.85, 1}
\newcommand{\todo}[1]{{\color{red}#1}}
\newcommand{\TODO}[1]{\textbf{\color{red}[TODO: #1]}}

\definecolor{cvprblue}{rgb}{0.21,0.49,0.74}
\usepackage[pagebackref,breaklinks,colorlinks,allcolors=cvprblue]{hyperref}
\definecolor{darkgreen}{rgb}{0.0, 0.5, 0.0}
\def\paperID{13661} 
\def\confName{CVPR}
\def\confYear{2025}

\title{Few-shot Personalized Scanpath Prediction}

\author{Ruoyu Xue\textsuperscript{1}, Jingyi Xu\textsuperscript{1}, Sounak Mondal\textsuperscript{1}, Hieu Le\textsuperscript{2},   Gregory Zelinsky\textsuperscript{1}, Minh Hoai\textsuperscript{3}, Dimitris Samaras\textsuperscript{1} \\
\textsuperscript{1}Stony Brook University, USA~~~  
\textsuperscript{2}EPFL, Switzerland~~~ 
\textsuperscript{3}The University of Adelaide, Australia
}

\begin{document}
\def\mA{\mathcal{A}}
\def\mB{\mathcal{B}}
\def\mC{\mathcal{C}}
\def\mD{\mathcal{D}}
\def\mE{\mathcal{E}}
\def\mF{\mathcal{F}}
\def\mG{\mathcal{G}}
\def\mH{\mathcal{H}}
\def\mI{\mathcal{I}}
\def\mJ{\mathcal{J}}
\def\mK{\mathcal{K}}
\def\mL{\mathcal{L}}
\def\mM{\mathcal{M}}
\def\mN{\mathcal{N}}
\def\mO{\mathcal{O}}
\def\mP{\mathcal{P}}
\def\mQ{\mathcal{Q}}
\def\mR{\mathcal{R}}
\def\mS{\mathcal{S}}
\def\mT{\mathcal{T}}
\def\mU{\mathcal{U}}
\def\mV{\mathcal{V}}
\def\mW{\mathcal{W}}
\def\mX{\mathcal{X}}
\def\mY{\mathcal{Y}}
\def\mZ{\mathcal{Z}} 

\def\bbN{\mathbb{N}} 
\def\bbR{\mathbb{R}} 
\def\bbP{\mathbb{P}} 
\def\bbQ{\mathbb{Q}} 
\def\bbE{\mathbb{E}}

\def\1n{\mathbf{1}_n}
\def\0{\mathbf{0}}
\def\1{\mathbf{1}}

\def\A{{\bf A}}
\def\B{{\bf B}}
\def\C{{\bf C}}
\def\D{{\bf D}}
\def\E{{\bf E}}
\def\F{{\bf F}}
\def\G{{\bf G}}
\def\H{{\bf H}}
\def\I{{\bf I}}
\def\J{{\bf J}}
\def\K{{\bf K}}
\def\L{{\bf L}}
\def\M{{\bf M}}
\def\N{{\bf N}}
\def\O{{\bf O}}
\def\P{{\bf P}}
\def\Q{{\bf Q}}
\def\R{{\bf R}}
\def\S{{\bf S}}
\def\T{{\bf T}}
\def\U{{\bf U}}
\def\V{{\bf V}}
\def\W{{\bf W}}
\def\X{{\bf X}}
\def\Y{{\bf Y}}
\def\Z{{\bf Z}}

\def\a{{\bf a}}
\def\b{{\bf b}}
\def\c{{\bf c}}
\def\d{{\bf d}}
\def\e{{\bf e}}
\def\f{{\bf f}}
\def\g{{\bf g}}
\def\h{{\bf h}}
\def\i{{\bf i}}
\def\j{{\bf j}}
\def\k{{\bf k}}
\def\l{{\bf l}}
\def\m{{\bf m}}
\def\n{{\bf n}}
\def\o{{\bf o}}
\def\p{{\bf p}}
\def\q{{\bf q}}
\def\r{{\bf r}}
\def\s{{\bf s}}
\def\t{{\bf t}}
\def\u{{\bf u}}
\def\v{{\bf v}}
\def\w{{\bf w}}
\def\x{{\bf x}}
\def\y{{\bf y}}
\def\z{{\bf z}}

\def\balpha{\mbox{\boldmath{$\alpha$}}}
\def\bbeta{\mbox{\boldmath{$\beta$}}}
\def\bdelta{\mbox{\boldmath{$\delta$}}}
\def\bgamma{\mbox{\boldmath{$\gamma$}}}
\def\blambda{\mbox{\boldmath{$\lambda$}}}
\def\bsigma{\mbox{\boldmath{$\sigma$}}}
\def\btheta{\mbox{\boldmath{$\theta$}}}
\def\bomega{\mbox{\boldmath{$\omega$}}}
\def\bxi{\mbox{\boldmath{$\xi$}}}
\def\bnu{\mbox{\boldmath{$\nu$}}}                                  
\def\bphi{\mbox{\boldmath{$\phi$}}}
\def\bmu{\mbox{\boldmath{$\mu$}}}

\def\bDelta{\mbox{\boldmath{$\Delta$}}}
\def\bOmega{\mbox{\boldmath{$\Omega$}}}
\def\bPhi{\mbox{\boldmath{$\Phi$}}}
\def\bLambda{\mbox{\boldmath{$\Lambda$}}}
\def\bSigma{\mbox{\boldmath{$\Sigma$}}}
\def\bGamma{\mbox{\boldmath{$\Gamma$}}}
                                  
\newcommand{\myprob}[1]{\mathop{\mathbb{P}}_{#1}}

\newcommand{\myexp}[1]{\mathop{\mathbb{E}}_{#1}}

\newcommand{\mydelta}[1]{1_{#1}}

\newcommand{\myminimum}[1]{\mathop{\textrm{minimum}}_{#1}}
\newcommand{\mymaximum}[1]{\mathop{\textrm{maximum}}_{#1}}    
\newcommand{\mymin}[1]{\mathop{\textrm{minimize}}_{#1}}
\newcommand{\mymax}[1]{\mathop{\textrm{maximize}}_{#1}}
\newcommand{\mymins}[1]{\mathop{\textrm{min.}}_{#1}}
\newcommand{\mymaxs}[1]{\mathop{\textrm{max.}}_{#1}}  
\newcommand{\myargmin}[1]{\mathop{\textrm{argmin}}_{#1}} 
\newcommand{\myargmax}[1]{\mathop{\textrm{argmax}}_{#1}} 
\newcommand{\myst}{\textrm{s.t. }}

\newcommand{\denselist}{\itemsep -1pt}
\newcommand{\sparselist}{\itemsep 1pt}

\definecolor{pink}{rgb}{0.9,0.5,0.5}
\definecolor{purple}{rgb}{0.5, 0.4, 0.8}   
\definecolor{gray}{rgb}{0.3, 0.3, 0.3}
\definecolor{mygreen}{rgb}{0.2, 0.6, 0.2}

\newcommand{\cyan}[1]{\textcolor{cyan}{#1}}
\newcommand{\blue}[1]{\textcolor{blue}{#1}}
\newcommand{\magenta}[1]{\textcolor{magenta}{#1}}
\newcommand{\pink}[1]{\textcolor{pink}{#1}}
\newcommand{\green}[1]{\textcolor{green}{#1}} 
\newcommand{\gray}[1]{\textcolor{gray}{#1}}    
\newcommand{\mygreen}[1]{\textcolor{mygreen}{#1}}    
\newcommand{\purple}[1]{\textcolor{purple}{#1}}       

\definecolor{greena}{rgb}{0.4, 0.5, 0.1}
\newcommand{\greena}[1]{\textcolor{greena}{#1}}

\definecolor{bluea}{rgb}{0, 0.4, 0.6}
\newcommand{\bluea}[1]{\textcolor{bluea}{#1}}
\definecolor{reda}{rgb}{0.6, 0.2, 0.1}
\newcommand{\reda}[1]{\textcolor{reda}{#1}}

\def\changemargin#1#2{\list{}{\rightmargin#2\leftmargin#1}\item[]}
\let\endchangemargin=\endlist
                                               
\newcommand{\cm}[1]{}

\newcommand{\mhoai}[1]{{\color{blue}\textbf{[MH: #1]}}}

\newcommand{\mtodo}[1]{{\color{red}$\blacksquare$\textbf{[TODO: #1]}}}
\newcommand{\myheading}[1]{\vspace{0.5ex}\noindent \textbf{#1}}
\newcommand{\htimesw}[2]{\mbox{$#1$$\times$$#2$}}


%
%
%

\newcommand{\Sref}[1]{Sec.~\ref{#1}}
\newcommand{\Eref}[1]{Eq.~(\ref{#1})}
\newcommand{\Fref}[1]{Fig.~\ref{#1}}
\newcommand{\Tref}[1]{Table~\ref{#1}}

\newcolumntype{C}[1]{>{\centering\arraybackslash}p{#1}}
\maketitle
\renewcommand{\thefootnote}{\fnsymbol{footnote}}
\newif\ifdraft
\drafttrue 
\definecolor{darkpink}{rgb}{0.91, 0.33, 0.5}
\ifdraft
  \newcommand{\HL}[1]{{\color{orange}{\bf HL: #1}}} 
 \newcommand{\hl}[1]{{\color{orange} #1}}
  \newcommand{\JX}[1]{{\color{darkpink}{\bf JX: #1}}} 
 \newcommand{\jx}[1]{{\color{darkpink} #1}}
 
  \newcommand{\ME}[1]{{\color{darkgreen}{\bf ME: #1}}}
 \newcommand{\me}[1]{{\color{darkgreen} #1}}

\else
  \newcommand{\HL}[1]{}
 \newcommand{\hl}[1]{#1}
 \newcommand{\ME}[1]{}
  \newcommand{\me}[1]{#1}
  \newcommand{\TODO}[1]{}
  \newcommand{\todo}[1]{#1}
\fi

\newcommand{\parag}[1]{\vspace{-3mm}\paragraph{#1}}
\begin{abstract}

A personalized model for scanpath prediction provides insights into the visual preferences and attention patterns of individual subjects. However, existing methods for training scanpath prediction models are data-intensive and cannot be effectively personalized to new individuals with only a few available examples. In this paper, we propose few-shot personalized scanpath prediction task (FS-PSP) and a novel method to address it, which aims to predict scanpaths for an unseen subject using minimal support data of that subject's scanpath behavior. The key to our method's adaptability is the Subject-Embedding Network (SE-Net), specifically designed to capture unique, individualized representations for each subject's scanpaths. SE-Net generates subject embeddings that effectively distinguish between subjects while minimizing variability among scanpaths from the same individual. The personalized scanpath prediction model is then conditioned on these subject embeddings to produce accurate, personalized results. Experiments on multiple eye-tracking datasets demonstrate that our method excels in FS-PSP settings and does not require any fine-tuning steps at test time. Code is available at: \href{https://github.com/cvlab-stonybrook/few-shot-scanpath}{https://github.com/cvlab-stonybrook/few-shot-scanpath}


\end{abstract}    
\section{Introduction}
\label{sec:intro}

Recent models of scanpath prediction have excelled at predicting human attention \cite{chen2021predicting, chen2024beyond, yang2020predicting, chen2022characterizing, yang2022target, mondal2023gazeformer, mondal2025look, yang2024unifying}, which is important for applications such as autonomous driving~\cite{chung2022static, navarro2021dynamic}, virtual and augmented reality~\cite{kapp2021arett, sui2023scandmm}, healthcare diagnostics~\cite{carette2019learning, song2024ems}, and information visualization~\cite{wang2023scanpath}. However, these methods are trained using attention data from multiple subjects and therefore learn population-level ``average'' attention patterns that will fail to reflect individual differences shaped by culture, memory, and experience \cite{judd2012benchmark}. To capture these individualized attention patterns, and to avoid biases that can result from group averaging, models of personalized scanpath prediction (PSP) attempt to learn individual subject embeddings that are used to predict an individual's scanpath~\cite{chen2024beyond, yuan2023personalized, jiang2024eyeformer}. PSP is particularly useful for applications such as recommendation systems~\cite{shahriar2020online, chen2023eye, song2019eye} and advertisements~\cite{pfiffelmann2020personalized} because it allows subject personality to be decoded from an individual's attention.   
\begin{figure}[t]
  \centering
  \includegraphics[trim={0 0 0.3cm 0},clip,width=1.0\linewidth]{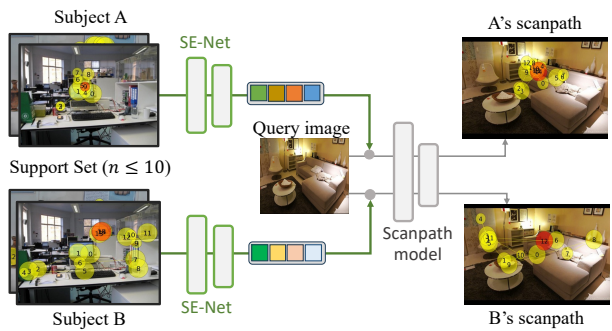}
  \caption{\textbf{Few-shot Personalized Scanpath Prediction (FS-PSP)}. Given a new subject with only a few support examples of their gaze behavior, can we adapt a base scanpath prediction model to this subject? We propose a subject-embedding extracting network, SE-Net, to achieve this personalized adaption.
  }
  \label{fig:teaser}
\end{figure}


A limitation of existing PSP models is that they require extensive data to accurately capture individual attention patterns. For example, to train models to predict the scanpaths made by people as they are searching for objects \cite{yang2020predicting}, it was necessary to collect the COCO-Search18 dataset of 269,760 search fixations \cite{chen2021predicting}, an effort requiring 10 subjects to each come to a laboratory and have their eye movements recorded for 10--12 hours. To be useful in practice, PSP models must be trainable on orders of magnitude less data. We therefore introduce the task of \textit{few-shot personalized scanpath prediction} (FS-PSP) where an individual's attention must be predicted using fixations from only a few behavioral observations, defined here as the scanpaths of people viewing $\leq 10$ images. We refer to the image-scanpath pairs collected for each subject as \textit{support samples}. 

FS-PSP is challenging because individuals' patterns of attention must be captured in just a few support samples. 
Existing PSP methods~\cite{chen2024beyond, yuan2023personalized, jiang2024eyeformer} often overfit to limited image content due to their reliance on large training datasets and a lack of efficient adaptation mechanisms. Typically, the primary goal of these methods is to jointly learn a subject embedding along with scanpath patterns to improve performance on ``\textit{seen}'' subjects, \textit{i.e.}, ones with sufficient training scanpath data. However, subject embeddings are treated as a byproduct of scanpath prediction, limiting these models' ability to leverage knowledge from previously seen embeddings to adapt to new subjects. Consequently, overfitting on a few scenes and insufficient representation of novel individual visual patterns both cause PSP model performance to drop substantially in few-shot settings, as in the case of ISP~\cite{chen2024beyond} with ten support samples. EyeFormer\cite{jiang2024eyeformer} also requires a support set with at least 50 scanpaths to achieve stable personalized embeddings.

In this paper, we propose a flexible scanpath model that can adapt to new subject without requiring retraining or fine-tuning. This is achieved by decoupling subject embedding learning from the scanpath prediction process. First, we learn a subject embedding space that encodes personalized attention traits, then condition the scanpath prediction model on these embeddings. This separation enables robust performance in few-shot scenarios by avoiding the complexities of joint learning: the subject embedding extractor focuses exclusively on capturing the unique features of each subject's scanpath, while the scanpath prediction model only needs to learn a conditional mapping based on the corresponding subject embedding.

More specifically, we propose a Subject Embedding Network (SE-Net) to extract subject embeddings from image-scanpath pairs. During training, we use a base dataset containing a large number of images and scanpaths collected from the seen subjects. SE-Net is trained with a classification loss to distinguish between these subjects, focusing on extracting distinctive, personalized traits that support effective personalization. Additionally, we apply a contrastive loss to ensure embeddings from different scanpaths of the same subject remain similar while being distinct across different subjects. This training strategy emphasizes extracting robust, unbiased, and representative embeddings for each subject, akin to previous representation-learning-based few-shot approaches~\cite{Yang2022FewShotCW,jian-etal-2022-contrastive,xu2022generating,xu2021variational}. We then train a personalized scanpath prediction network~\cite{chen2024beyond} on the base dataset using these learned subject embeddings, enhancing its capacity to infer scanpaths based on the given subject embeddings. At the inference stage, we extract embeddings for unseen subjects from a few image-scanpath exemplars (support set), average them to obtain a single subject embedding, and condition the scanpath prediction model on this embedding to enable generalization to new subjects.

To demonstrate the performance of our method on FS-PSP, we predict personalized scanpaths of unseen subjects under three $n$-shot settings ($n=1,5,10$) and on three datasets: OSIE~\cite{xu2014predicting}, COCO-Freeview~\cite{chen2022characterizing}, and COCO-Search18~\cite{yang2020predicting}. 
Our method outperforms the second-best model on the ScanMatch metric\cite{cristino2010scanmatch} on these datasets by 5.9\%, 7.9\% and 6.0\%.


\section{Related Works}
\textbf{Personalied Scanpath Prediction}. Scanpath prediction aims to model both bottom-up and top-down human attention by predicting sequences of fixations as observers either freely explore a scene without specific instructions\cite{kummerer2022deepgaze, wang2023scanpath, sui2023scandmm, assens2017saltinet, assens2018pathgan, sun2019visual}, or engage in task-specific activities, such as visual question answering\cite{chen2021predicting}, webpage browsing, searching\cite{mondal2023gazeformer, chen2021coco, yang2020predicting, yang2022target, yang2024unifying}, and object referral\cite{mondal2025look}. Personalized scanpath prediction (PSP) aims to predict the scanpath of an individual subject rather than the group average. Although there are multiple works on personalized saliency prediction\cite{li2018personalized, chang2019salgaze, xu2018personalized, chen2023learning, strohm2024learning}, PSP is a relatively new and underexplored task, as it requires building upon robust population-level scanpath prediction models. The key challenge in personalized scanpath prediction is enabling the model to recognize which subject's scanpath it is predicting. Current approaches \cite{chen2024beyond,jiang2024eyeformer,yuan2023personalized} incorporate subject embeddings to represent different individuals. ISP \cite{chen2024beyond} adds three modules to existing scanpath prediction models \cite{mondal2023gazeformer, chen2021predicting} to encode subject embeddings at different stages within the model. EyeFormer\cite{jiang2024eyeformer} designs a new scanpath prediction model that leverages reinforcement learning and uses a viewer encoder to indicate different subjects. However, ISP suffers significant performance drops when working with very limited data ($N\leq10$) for a new subject, while EyeFormer has only demonstrated its effectiveness with no fewer than 50 samples. This issue arises because over-parameterized models tend to overfit on small datasets, and must relearn subject embedding for the new subject, limiting their ability to fully utilize prior experience with previously seen subjects. To address these issues, we propose SE-Net, designed to mitigate overfitting and effectively leverage model experience.\\
\textbf{Few-shot Learning Leveraging User Emebedding.} Few-shot learning (FSL) has been widely studied in various topics, and a commonly used approach in FSL is to learn a function that maps the input to an embedding to effectively represent the prototype of different classes\cite{ye2020few, han2022meta, snell2017prototypical, xu2022generating, xu2023generating, xu2021variational, li2021adaptive, wang2019panet}. User embedding, \ie embeddings to represent person-specific patterns, encodes a user's unique behavioral patterns by mapping their actions through a network, capturing both intra-personal and inter-personal similarities and differences. This not only enables downstream tasks to effectively distinguish between individuals but, more importantly, allows for an understanding of unique user traits from the user embedding, enabling predictions that align with each person’s distinctive patterns. Many topics address the few-shot problem by learning user embeddings, including recommendation systems~\cite{li2020few} and gaze estimation models~\cite{park2019few, he2019device}. Although several works utilize scanpath for classification\cite{zhong2024spformer, nishiyasu2024gaze}, to the best of our knowledge, this is the first work that aims to extract user embeddings from scanpaths—a non-trivial task requiring the disentanglement of visual patterns driven by both bottom-up and top-down attention from fixation locations, temporal sequences, and durations.
\section{Proposed FS-PSP Framework}
\begin{figure*}[t]
  \centering
  \includegraphics[width=1.0\linewidth]{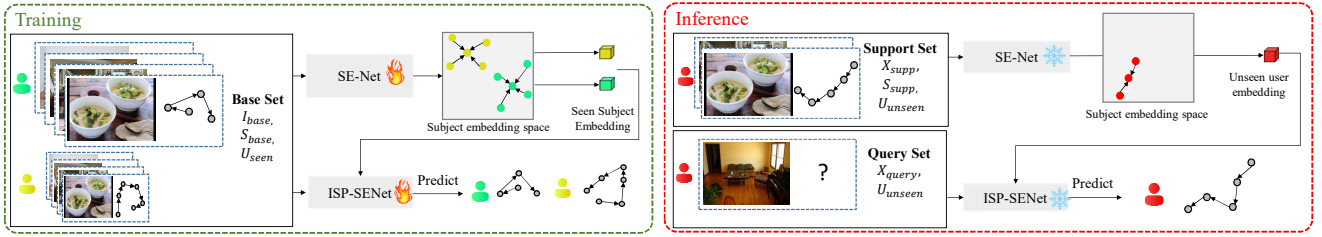}
  \caption{\textbf{Overview of ISP-SENet}: Our method for few-shot personalized scanpath prediction has two stages. In the training stage, we train two models on a large amount of image-scanpath pairs $\mD_{base}$, corresponding to a set of seen subjects. Initially, we train the Subject Embedding Network (SE-Net) to obtain embeddings for seen subjects, followed by training ISP-SENet to predict scanpaths using these embeddings. In the inference phase, both models are frozen, and we extract embeddings for unseen subjects from the support set, $\mD_{supp}$, which consists of $n$-shot images sampled from the base set. These unseen subject embeddings then guide ISP-SENet in predicting scanpaths for unseen subjects using the query set, $\mD_{query}$, which includes a collection of unseen images.}
  \label{fig:overview}
\end{figure*}
This section describes our framework for FS-PSP, which consists of two major components: (1) a personalized scanpath predictor, called ISP-SENet, that predicts the scanpath for a subject, conditioned on the subject's  embedding, and (2) a subject-embedding network, called SE-Net, that computes the embedding for a subject based on a small set of their gaze behaviors. In this section, we first provide a formal definition of the task and then describe the different components of our framework. The overview is shown in \cref{fig:overview}.

\subsection{Problem Formulation}

Let $\d$ represent a scanpath data instance; it is a tuple consisting of an image and a sequence of 2D gaze fixations. Let $S(\d)$ denote the identity of the subject from whom the scanpath was collected. PSP considers a scenario where we have a base training set of gaze behavior with size $m$: $\mD_{base} = \{\d_{base}^i\}_{i=1}^{m}$, which can be used to train a personalized scanpath predictor, but only for the subjects in the base training set. FS-PSP goes beyond PSP, considering the task of personalized scanpath prediction for a subject $\s$ who was not seen during training ($\s  \neq S(\d_{base}^i) \ \forall i$) but for whom we have some gaze behavior data on a small subset of images from the base dataset, referred to as the support set. Let $\mD_{supp} = \{\d_{supp}^i\}_{i=1}^{n}$ represent this support set, where $S(\d_{supp}^i) = \s \  \forall i$, and $n$ is small (at most 10). The goal of FS-PSP is, given a set of unseen subjects, to predict their scanpaths on a set of query images $\x^i$ with size $q$: $\mD_{query} = \{\x_{query}^i\}_{i=1}^{q}$.

\subsection{ISP-SENet -- Personalized Scanpath Predictor}

ISP-SENet is one of the two core components of our framework. In this paper, we propose developing it based on Gazeformer-ISP \cite{chen2024beyond}. This model accounts for the scanpath behaviors of individual subjects, with each subject in the training set associated with a separate embedding vector that is learnable but fixed after training. By assigning a unique embedding vector to each subject, the model achieves significant improvement over a generic model that does not account for subject identity. However, this approach only works for subjects seen during training, as the embedding vector can be retrieved from a lookup table. For an unseen subject, there is no way to compute the embedding vector, thereby preventing prediction for new subjects. 

In this work, we propose replacing the fixed embedding vector with the output of a subject-embedding network, SE-Net. SE-Net can compute the subject embedding vector for any subject, as long as a support set of gaze behavior data from that subject is available. SE-Net represents the main technical innovation of our work, which we describe in the following section.

\subsection{SE-Net -- Subject Embedding Network}


SE-Net computes a function $f$ that takes a scanpath (both an image and a trajectory of eye fixations with durations) as input and outputs an embedding vector. This network calculates the subject embedding given a single scanpath behavior data point. For a support set contains more than one gaze behavior data point, $\mD_{supp} = \{\d_{supp}^i\}_{i=1}^{n}$, we follow the approach of prototypical networks~\cite{snell2017prototypical} and simply take the average as the overall subject's embedding, i.e., $\frac{1}{n}\sum_{i=1}^{n} f(\d^i_{supp})$. We will next describe how this network is trained and then provide details about its architecture.

\subsubsection{Training SE-Net}
\label{sec:training}
\begin{figure*}[t!]
  \centering
  \includegraphics[width=0.8\linewidth]{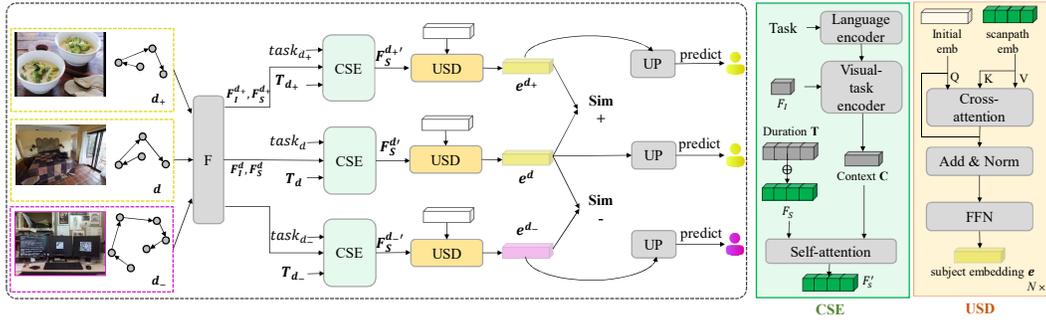}
  \caption{\textbf{Structure of SE-Net.} SE-Net employs a feature extractor $F$ to derive image and scanpath semantic features, $F_I$ and $F_S$, respectively. The CSE module then processes task and duration features, updating the scanpath embedding constrained by all extracted features. An initialized embedding learns human attention information from $F_S'$ to produce the subject embedding $e$. This triplet network assesses the distances among $e^{d_+}, e^{d_+}, e^{d_-}$, and the UP module predicts the subject ID. All CSE modules share the same weights, as do the USD and UP modules.}
  \label{fig:senet}
\end{figure*}


SE-Net is trained using scanpath data from the base training set $\mD_{\text{base}}$, starting with the creation of triplets $(\d, \d_+, \d_-)$, where $\d$ is a scanpath drawn from $\mD_{\text{base}}$, $\d_+$ is another scanpath randomly drawn from $\mD_{\text{base}}$ but from the same subject as $\d$, and $\d_-$ is a scanpath randomly drawn from a different subject. The training loss for this triplet is the combination of three classification losses and a contrastive loss term: 
\begin{align}
	\mL_{cls}(\d) + \mL_{cls}(\d_+) + \mL_{cls}(\d_-) + \mL_{\text{contrast}}. 
\end{align}
The classification loss $\mathcal{L}_{\text{cls}}$ is computed separately for each of $\d$, $\d+$, and $\d_-$. It is defined as classifying the subject given their scanpath on an image by adding a classification head to the subject embedding layer. 
%
%
The contrastive loss $\mathcal{L}_{\text{contrast}}$ is based on triplet loss \cite{schroff2015facenet}:
\begin{align}
\max \left( \| f(\d) - f(\d_+) \|^2 - \| f(\d) - f(\d_-) \|^2 + m, 0 \right).
\label{eq:eq8}
\end{align}
The margin $m$ controls the distance between subject embeddings. A smaller margin allows scanpaths from different subjects to be similar, typically observed in top-down attention scenarios. Conversely, a larger margin is preferable in scenarios like free-viewing, characterized by significant diversity among subjects. Refer to the supplementary for more detailed explanations.

%


\subsubsection{Network architecture}
\label{sec:network-architecture}
\myheading{Feature extraction.} Our semantic feature extractor $F$ follows the design of HAT~\cite{yang2024unifying}. We encode images and scanpaths using hierarchical feature maps from an image encoder (ResNet \cite{he2016deep}) and decoder (Deformable attention \cite{zhu2020deformable}) and obtain image tokens $F_I \in \mathbb{R}^{\left(\frac{H}{32} \cdot \frac{W}{32}\right) \times C}$ and scanpath tokens $F_S \in \mathbb{R}^{L \times c}$ ($H, W$ are the height and width of image, $L$ is the max length of scanpaths, $c$ is embedding dimension). The details are described in the supplementary material. \\

\myheading{Context-Scanpath Encoder(CSE).} Human attention can be divided into bottom-up and top-down processes, either freely viewing the image or viewing it with a specific task, such as searching for a target. Different tasks influence attention toward specific objects mentioned in the task and related objects. To make SE-Net task-aware, we design an multi-modal visual-task encoder for SE-Net, which takes task and image tokens as input, highlighting the image content relevant to the task. The task is first mapped to a task embedding $t \in \mathbb{R}^{c}$ by a language model (RoBERTa \cite{liu2019roberta}), then the visual-task encoder is designed as \cref{eq:eq1}:
\begin{align}
    C = \text{SelfAttn}(\{t \cdot W, F_I\}).
    \label{eq:eq1}
\end{align}
Here, $C$ represents the context embedding, capturing information from both the image and the task. $W$ is a linear layer. $\{ , \}$ denotes concatenation along the sequence dimension, and $\text{SelfAttn}$ is multiple self-attention layers. \\
To obtain the final scanpath embedding, we first ensure the model is aware of each fixation's duration. Position information reflects where the subject tends to focus, while duration indicates the relative interest in specific objects. We use 2D sinusoidal position embeddings \cite{li2021learnable, yang2024unifying} to encode the position information and 1D positional embeddings for durations. Given that durations range from 0 to 5000 ms, using 5000 different positional encodings would be redundant, as small differences (\eg, 200 ms vs. 203 ms) are often negligible. Therefore, we collected all durations in the base set and uniformly grouped them into 10 bins, replacing the actual duration with its corresponding bin ID. This replacement is only used in SE-Net, while ISP-SENet still predicts the raw duration. Find details in the supplementary. \\
Combining the context embedding $C$, duration embeddings $\mathcal{T}$, and position embeddings $Pos$, we update the scanpath tokens $F_S'$ representing viewing behavior refined by image content and task as shown in \cref{eq:eq2}, w/o $C$ denotes that tokens associated with $C$ are discarded, Isolating image content from subject embeddings to reduce scene-specific bias.
\begin{equation}
    F_S' = \text{SelfAttn}(\{C, F_S + \mathcal{T} + \text{Pos}\})_{\text{w/o } C}
    \label{eq:eq2}
\end{equation}
\myheading{User-Scanpath Decoder (USD).} The main focus of USD is to capture subject attention traits from the scanpath embedding. We initialize a subject token $e \in \mathbb{R}^{c}$ and use a cross-attention mechanism to extract the subject embedding from the scanpath embedding. Here, the subject token serves as the query, enabling it to attend to the scanpath embedding (used as key and value) and extract specific patterns that reflect the subject's unique attention characteristics. The detailed implementation is shown in \cref{eq:eq3}:
\begin{equation}
    e = \text{ReLU}(\text{Linear}(e + \text{CrossAttn}(e, F_S'))).
    \label{eq:eq3}
\end{equation}
To enable the updated subject embedding to distinguish between different subjects, we add a subject Predictor (UP), a series of linear layers with ReLU activation, to classify the subject ID:
\begin{equation}
   \text{Subject ID}= \text{Linear}(\text{ReLU}(\text{Linear}(e))
   \label{eq:eq4}
\end{equation}

\section{Experiment}
\subsection{Dataset Setting}

\myheading{Dataset}. We train and evaluate ISP-SENet on three datasets: OSIE~\cite{xu2014predicting}, COCO-Freeview~\cite{chen2022characterizing}, and COCO-Search18~\cite{yang2020predicting} (target-present). OSIE collects scanpaths under a free-viewing condition from 15 subjects on 700 images. COCO-Freeview collects scanpaths under a free-viewing condition from 10 subjects on 6202 images. COCO-Search18 has same images as COCO-FreeView, but the scanpaths are collected under a search task on 18 categories. We follow the same data split of OSIE as ISP~\cite{chen2024beyond}, and the same split of COCO-Search18 and COCO-FreeView as HAT\cite{yang2024unifying}.

\myheading{Unseen subject selection}. For each dataset, we split the full subject set into around 70\% seen and 30\% unseen. We repeat experiments twice with different splits, and the result of another split is shown in the supplementary.

\myheading{n-shot sampling}. We selected three values, $n=1,5,10$, as criteria for choosing the support set. For each value of $n$, we randomly sample $n$ image-scanpath pairs from unseen subjects within the training set and assessed the model's performance across the entire test set of unseen subjects. This process was repeated 10 times with various support sets to ensure robustness, and the average performance metrics were reported. Find margin of error in the supplementary.

\subsection{Experiment Setting}

\myheading{Baselines}. We use ChenLSTM-ISP and Gazeformer-ISP\cite{chen2024beyond} as baselines designed for personalized scanpath prediction. Both models are pretrained on a base set and fully fine-tuned on a support set to jointly learn subject embeddings and scanpaths. Since new subject embeddings are not available initially, fine-tuning is required. To reduce overfitting, we also implement ChenLSTM-ISP-S and Gazeformer-ISP-S, which only fine-tune the subject embedding layer and freeze all other parameters.

\myheading{Metrics}. We evaluated ISP-SENet using value-based metrics as outlined by ISP~\cite{chen2024beyond}, including ScanMatch (SM), MultiMatch (MM), and String-Edit Distance (SED), assess the similarity between predicted and actual scanpaths. ScanMatch~\cite{cristino2010scanmatch} evaluates the overall similarity, MultiMatch~\cite{anderson2015comparison, dewhurst2012depends} analyzes five dimensions of scanpath similarity: shape, direction, length, position, and duration, while String-Edit Distance~\cite{brandt1997spontaneous} identifies structural differences. 

\begin{table*}[t]
    \setlength{\intextsep}{10mm}
    \centering
    \renewcommand{\arraystretch}{1.0} 
    \setlength{\tabcolsep}{3pt} 
     \resizebox{1.6\columnwidth}{!}{
    \begin{tabular}{c|c|ccc|ccc|ccc}
        \multirow{2}{*}{$n$-shot} & \multirow{2}{*}{\textbf{Method}} & \multicolumn{3}{c|}{\textbf{OSIE}} & \multicolumn{3}{c|}{\textbf{COCO-FreeView}} & \multicolumn{3}{c}{\textbf{COCO-Search18}} \\
        & & SM $\uparrow$ & MM $\uparrow$ & SED $\downarrow$ & SM $\uparrow$ & MM $\uparrow$ & SED $\downarrow$ & SM $\uparrow$ & MM $\uparrow$ & SED $\downarrow$ \\
        \midrule

        \multirow{2}{*}{$n=1$}
        & ChenLSTM-ISP & 0.282 & 0.763 & 7.832 & 0.287 & 0.805 & 13.307 & 0.371 & 0.760 & 2.756 \\
        & Gazeformer-ISP & 0.327 & 0.792 & 7.873 & 0.244 & 0.787 & 15.118 & 0.342 & 0.770 & 2.818 \\
        & ChenLSTM-ISP-S & 0.328 & 0.793 & 7.601 & 0.339 & 0.814 & 12.523 & 0.448 & 0.803 & 2.394 \\
        & Gazeformer-ISP-S & 0.354 & 0.801 & 7.503 & 0.333 & 0.817 & 12.538 & 0.446 & 0.802 & 2.463 \\
        & \textbf{ISP-SENet} & \textbf{0.368} & \textbf{0.805} & \textbf{7.413} & \textbf{0.369} & \textbf{0.832} & \textbf{12.227} & \textbf{0.475} & \textbf{0.814} & \textbf{2.333} \\
        \midrule

        \multirow{2}{*}{$n=5$} 
        & ChenLSTM-ISP & 0.319 & 0.773 & 7.855 & 0.320 & 0.815 & 12.950 & 0.386 & 0.773 & 2.489 \\
        & Gazeformer-ISP & 0.340 & 0.791 & 7.920 & 0.286 & 0.800 & 14.630 & 0.353 & 0.774 & 2.980 \\
        & ChenLSTM-ISP-S & 0.329 & 0.791 & 7.649 & 0.338 & 0.814 & 12.540 & 0.449 & 0.803 & 2.380 \\
        & Gazeformer-ISP-S & 0.354 & 0.801 & 7.499 & 0.333 & 0.817 & 12.539 & 0.445 & 0.803  & 2.457 \\
        & \textbf{ISP-SENet} & \textbf{0.376} & \textbf{0.803} & \textbf{7.337} & \textbf{0.368} & \textbf{0.829} & \textbf{12.017} & \textbf{0.484} & \textbf{0.815} & \textbf{2.354} \\
        \midrule

        \multirow{2}{*}{$n=10$}
        & ChenLSTM-ISP & 0.322 & 0.777 & 7.740 & 0.323 & 0.819 & 12.541 & 0.393 & 0.781 & 2.394 \\
        & Gazeformer-ISP & 0.345 & 0.794 & 7.916 & 0.317 & 0.805 & 14.224 & 0.370 & 0.785 & 2.765 \\
        & ChenLSTM-ISP-S & 0.328 & 0.791 & 7.637 & 0.340 & 0.814 & 12.532 & 0.449 & 0.803 & 2.379 \\
        & Gazeformer-ISP-S & 0.354 & 0.802 & 7.505 & 0.333 & 0.816 & 12.545 & 0.446 & 0.802  & 2.464 \\
        & \textbf{ISP-SENet} & \textbf{0.375} & \textbf{0.803} & \textbf{7.318} & \textbf{0.367} & \textbf{0.828} & \textbf{11.956} & \textbf{0.482} & \textbf{0.815} & \textbf{2.359} \\
        \midrule
    \end{tabular}}
    \vskip -0.1in
\caption{Performance Comparison on Different Datasets under different few-shot settings for FS-PSP. \label{tab:main-result}} 
\vspace{-5pt}
\end{table*}

\myheading{Implementation details}. For SE-Net, the number of transformer layers in visual-task encoder and self-attention of CSE, and the number of USD layers are set to 3. The embedding dimension is set to 384. We use AdamW~\cite{loshchilov2017decoupled} with learning rate 0.0001. We train SE-Net for 25 epochs with a batch size of 16. The max length of scanpaths in OSIE and COCO-FreeView is set to 20, and the max length in COCO-Search18 is set to 10. For ISP-SENet, adhering to the GazeformerISP~\cite{chen2024beyond} configuration, we matched the maximum scanpath lengths and embedding dimensions with SE-Net, retaining other hyperparameters. The fine-tuning on support set takes 10 epochs for supervised training and 10 epochs for self-critical sequence training (SCST)~\cite{chen2021predicting}.

\subsection{Main Results}
\label{sec:main-result}
We compared ISP-SENet with four baselines across three datasets as shown in \cref{tab:main-result}. The consistent performance across $n=1,5,10$ indicates ISP-SENet's effectiveness in minimizing the impact of individual image instances, allowing embeddings to focus on attention patterns rather than being biased by the support set.

\myheading{Results on OSIE.} ISP-SENet outperforms the second-best approach by 5.9\% and 2.5\% in SM and SED, respectively. It maintains stable performance in the 5-shot and 10-shot settings and achieves a competitive SM score of 0.368 with just one image-scanpath pair. 

\myheading{Results on COCO-FreeView.} Both ISP methods exhibit severe overfitting, particularly as COCO-FreeView is a larger dataset with more complex scenes and scanpaths. Nonetheless, ISP-SENet shows substantial enhancements in all settings, outperforming the baseline by 7.9\% and 4.5\% in SM and SED. \\

\myheading{Results on COCO-Search18}. We achieved 6.0\% and 2.5\% on SM and SED compared with the second-best. The consistent performance of COCO-Search18 highlights ISP-SENet's capability in capturing varied attention patterns across different search tasks. This success is attributed to the visual-task encoder, detailed further in \cref{sec:analysis}. \\

We observed that in the fine-tuning stage of Gazeformer-ISP-S and ChenLSTM-ISP-S, the unseen subject embeddings change minimally from seen subject embeddings. This suggests that simple learnable embeddings (instead of SE-Net) lack adaptability to unseen subjects. Their relatively better performance compared with fine-tuning all parameters likely stems from the similarity between seen and unseen subjects.\\

\myheading{Adaptation time}. A significant achievement of ISP-SENet is its rapid adaptation time to new subjects. Adaptation time refers to the duration required to update the scanpath prediction model, enabling it to infer the scanpaths of new subjects. For baseline methods, adaptation time is equivalent to the duration of fine-tuning on a support set, which is 161 and 267 seconds for ChenLSTM-ISP and Gazeformer-ISP respectively. In contrast, ISP-SENet's adaptation time is simply the time taken to obtain subject embeddings from SE-Net, which is 3.62 seconds. ISP-SENet achieves nearly real-time adaptation, enhancing efficiency and making the method more practical for real-world applications. 

\begin{table}[h!]
    \centering
    \renewcommand{\arraystretch}{1} 
    \setlength{\tabcolsep}{2pt} 
     
     \resizebox{0.99\columnwidth}{!}{
    \begin{tabular}{lcccccc}
        \toprule
        \multirow{2}{*}{Dataset} & \multicolumn{3}{c|}{ISP-SENet-Seen} & \multicolumn{3}{c}{ISP-SENet-Unseen} \\
        \cmidrule(lr){2-4} \cmidrule(lr){5-7}
         & SM $\uparrow$ & MM $\uparrow$ & SED $\downarrow$ & SM $\uparrow$ & MM $\uparrow$ & SED $\downarrow$ \\
        \midrule
        OSIE & 0.390 & 0.812 & 7.309 & 0.375 & 0.803 & 7.318 \\
        COCO-FreeView & 0.401 & 0.841 & 11.581 & 0.367 & 0.828 & 11.956 \\
        COCO-Search18 & 0.492 & 0.815 & 2.156 & 0.482 & 0.815 & 2.359 \\
        \bottomrule
    \end{tabular}        
    }
    \vskip -0.1in
\caption{Performance Comparison. Our method fully trained on all subjects (ISP-SENet-Seen), and trained on seen subjects and adapt to unseen subjects under 10-shot setting (ISP-SENet-unseen). \label{tab:upper-bound}}     
\vspace{-5pt}
\end{table} 

To establish the upper bound of prediction performance on unseen subjects (ISP-SENet-Unseen), we fully trained ISP-SENet using all available training data and evaluated it on the same subjects designated as unseen in the few-shot setting. The results are presented in \cref{tab:upper-bound}. Remarkably, even without training on the unseen subjects, ISP-SENet's performance approaches the upper bound on OSIE and COCO-Search18. While ISP-SENet-Unseen underperforms compared to ISP-SENet-Seen on COCO-FreeView, it is trained on a considerably larger dataset of 43,143 scanpaths featuring more complex scenes and patterns. Even in more complex datasets, ISP-SENet can extract meaningful personalized embeddings from unseen subjects.

We employed scanpath accuracy, same as R@1 implemented in ~\cite{chen2024beyond}, to measure the accuracy of the predicted scanpath for a subject by determining if it is the most similar to the ground truth among all predictions made on the same image across different subjects, reflecting its ability to distinguish subjects. Distinguishing unseen from seen subjects is challenging due to the dataset's limited subject pool, potentially biasing the model towards known subjects. Nevertheless, our method showed considerable improvement in \cref{tab:rank-based}, illustrating that our unseen subject embeddings effectively discern among subjects despite the limited training subjects.

\begin{table}[h!]
    \centering
    \renewcommand{\arraystretch}{1} 
    \setlength{\tabcolsep}{3pt} 
    
    \resizebox{1.0\columnwidth}{!}{
    \begin{tabular}{lccc}
        \toprule
         &  COCO-FreeView & COCO-Search18 & OSIE \\
        \midrule
        ChenLSTM-ISP & 33.66 & 34.28 & 21.14 \\
        Gazeformer-ISP & 31.99 & 31.73 & 21.36\\
        ChenLSTM-ISP-S & 31.95 & 33.17 & 20.17 \\
        Gazeformer-ISP-S & 31.87 & 33.53 & 18.85\\
        ISP-SENet & \textbf{35.57} & \textbf{35.25} & \textbf{22.85} \\
        \bottomrule
    \end{tabular}
    }
    \vskip -0.1in
    \caption{Scanpath accuracy (higher is better) shows the model's ability to distinguish predicted scanpaths from different subjects.\label{tab:rank-based} } 
    
\end{table}

\begin{figure*}[h]
  \centering
  \includegraphics[width=1.0\linewidth]{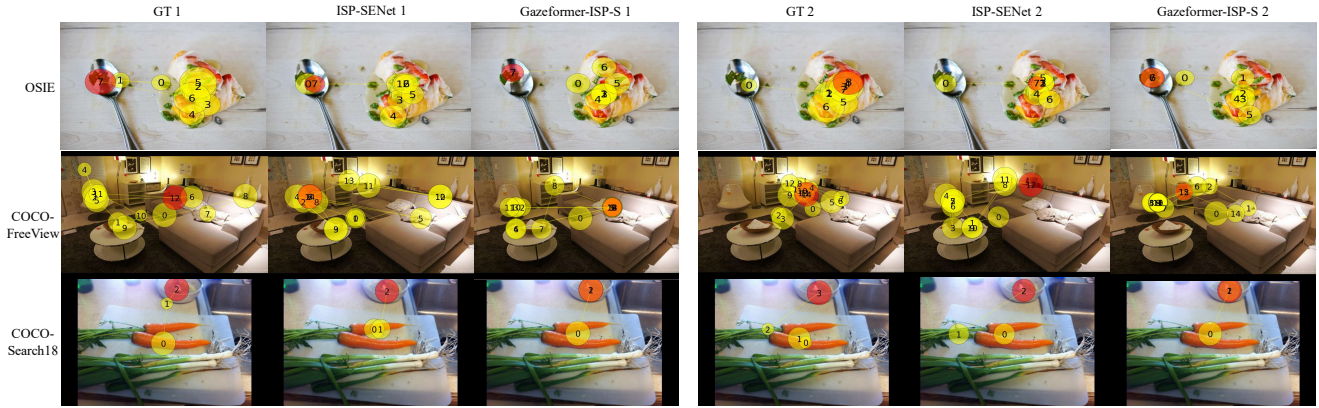}
  \caption{\textbf{Qualitative examples of scanpath prediction for different unseen subjects.} GT is the ground truth scanpaths of different unseen subjects. \textcolor{red}{Red} circle is the end fixation. In the third row, the subject is searching for ``bowl". The results indicates our method is able to capture the temporal order of fixations, fixation distributions, and distractions, while baseline keeps predicting similar scanpaths of different subjects.}
  \label{fig:main-vis}
\end{figure*}

\myheading{Qualitative results}. In \cref{fig:main-vis}, we showcase ISP-SENet's distinct scanpath predictions for the same query image across various unseen subjects, highlighting its ability to capture individual attention patterns. GT 1 and GT 2 represent ground truth scanpaths for different subjects. The first row shows ISP-SENet discerning varying fixation orders across objects. The second row demonstrates how ISP-SENet captures the diverse attention distributions: Subject 1 conducts a thorough scene scan, whereas Subject 2 focuses more centrally. The third row reveals Subject 2's distraction by peripheral objects during the search. These observations confirm ISP-SENet's effectiveness in identifying unique attention traits among unseen subjects for personalized scanpath prediction. Find more results in supplementary.

\section{Analysis}
\label{sec:analysis}
This section examines the impact of the number of seen subjects on performance, and the interpretability of the model. 

\begin{table}[h!]
    \centering
    \renewcommand{\arraystretch}{1} 
    \setlength{\tabcolsep}{4pt} 
    
    \resizebox{0.9\columnwidth}{!}{
    \begin{tabular}{lcccccc}
        \toprule
        \multirow{2}{*}{num seen} & \multicolumn{3}{c}{Subject 1} & \multicolumn{3}{c}{Subject 2} \\
        \cmidrule(lr){2-4} \cmidrule(lr){5-7}
         & SM $\uparrow$ & MM $\uparrow$ & SED $\downarrow$ & SM $\uparrow$ & MM $\uparrow$ & SED $\downarrow$ \\
        \midrule
        3 (20\%) & 0.339 & 0.806 & 7.871 & 0.354 & 0.781 & 6.740 \\
        7 (50\%) & 0.365 & 0.812 & 7.457 & 0.364 & 0.790 & 6.522 \\
        10 (67\%) & \textbf{0.374} & 0.814 & \textbf{7.420} & 0.385 & \textbf{0.794} & \textbf{6.460} \\
        13 (93\%) & 0.370 & \textbf{0.815} & 7.504 & \textbf{0.387} & 0.793 & 6.482 \\
        \bottomrule
    \end{tabular}
    }
    \vskip -0.1in
\caption{Performance Comparison of 10-shot setting on OSIE with different numbers of seen subjects in training stage. The table shows the performance for unseen subjects 1 and 2 with varying numbers of seen subjects. \label{tab:num-seen-subject} }         
\end{table}

\myheading{Size of the base training set.} Intuitively, training a model with a greater number of subjects enables it to learn a broader range of attention patterns. However, if trained on only a few subjects, the model might struggle with novel subjects whose scanpaths significantly deviate from those of seen subjects, thereby limiting its prediction accuracy. We assessed scanpath prediction performance for unseen subjects with varying numbers of training subjects, as detailed in \cref{tab:num-seen-subject}. We consistently treated two subjects (Subject 1 and 2 in the table) as unseen, sampling the support set ($n=10$) five times for each and averaging the results. Findings indicate that few-shot performance improves with an increasing number of seen subjects, although the difference between 10 and 13 subjects is minimal, likely due to performance saturation because of the potential similarity between some of OSIE dataset's subjects. See supplementary for COCO-Search18 results.

\begin{figure}[h]
  \centering
  \includegraphics[width=1\linewidth]{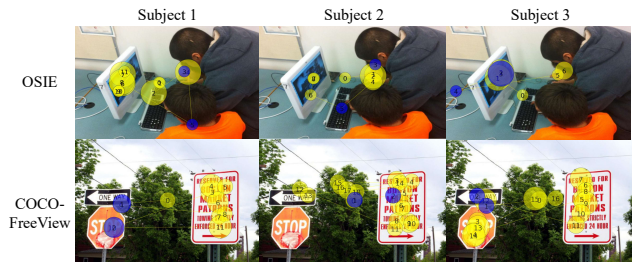}
  \caption{\textbf{Model interpretability.} By analyzing a large dataset of seen subject-scanpath pairs, SE-Net determines the most influential fixations in shaping unseen subject embeddings. Fixations highlighted in \textcolor{blue}{blue} represent the two with the highest weights. This analysis demonstrates that ISP-SENet can effectively identify which fixations are crucial for distinguishing between subjects.}
  \label{fig:model-interpretability}
\end{figure}

\myheading{Interpretability  -- which fixations reflect attention trait?} SE-Net also functions as a tool for quantitatively analyzing individual visual patterns. By examining the attention weights from the last cross-attention layer of the USD module, we can identify which fixation tokens the subject embeddings consider most critical. We illustrate this analysis in \cref{fig:model-interpretability}, showcasing the two fixations with the highest attention weights for each subject. For example, in the first row, the highlighted fixations focus on a kid, keyboard, and cables, which are key to distinguishing the three subjects' attention distributions. The second row captures the initial focus of three individuals on a stop sign, warning tag, and one-way tag in a traffic setting, hinting at SE-Net's potential to further study varied driving behaviors across different subjects.

\begin{figure}[h]
  \centering
  \includegraphics[width=1.0\linewidth]{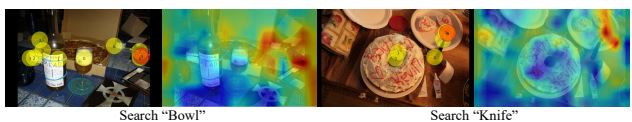}
  \caption{\textbf{Correspondence between task and image}. The visualization highlights the areas of the image that the visual-task encoder focuses on. The first and third columns display the attention weights extracted from the visual-task encoder, and the second and fourth columns show the ground truth scanpath visualizations from the support set. Fixations marked in \textcolor{red}{red} represent the last fixation, indicating the search target's location. This indicates the visual-task encoder's effectiveness in identifying task-relevant information within the image.}
  \label{fig:task-encoder}
\end{figure}

\myheading{Role of visual-task encoder.} COCO-Search18 features 18 search tasks, with participants typically focusing their final fixations on the search target in each image and task. The visual-task encoder informs the model about task-relevant image areas, providing prior knowledge of the search target for the CSE module to adjust its scanpath embedding. Although not tailored for object detection, the task embeddings consistently direct attention to target areas, as shown in the visualizations in \cref{fig:task-encoder}. See supplementary for quantitative results.

        

\begin{table}[h!]
    \centering
    \renewcommand{\arraystretch}{1} 
    \setlength{\tabcolsep}{4pt} 
    
    \resizebox{0.9\columnwidth}{!}{
    \begin{tabular}{lcccccc}
        \toprule
        \multirow{2}{*}{loss} & \multicolumn{3}{c}{OSIE} & \multicolumn{3}{c}{COCO-Search18} \\
        \cmidrule(lr){2-4} \cmidrule(lr){5-7}
         & SM $\uparrow$ & MM $\uparrow$ & SED $\downarrow$ & SM $\uparrow$ & MM $\uparrow$ & SED $\downarrow$ \\
        \midrule
        cls & 0.361 & 0.801 & 7.546 & 0.448 & 0.812 & 2.621 \\
        contrast & 0.372 & 0.795 & 7.389 & 0.442 & 0.812 & 2.400 \\
        cls+contrast & 0.375 & 0.803 & 7.318 & 0.482 & 0.815 & 2.359 \\
        \bottomrule
    \end{tabular}
    }
    \vskip -0.1in
\caption{Ablation on the effect of different losses. cls is solely using classification loss, contrast is solely using contrastive loss, cls+contrast combines both loss, consistent with the settings used in all main experiments. \label{tab:abalation-loss} }         
\end{table}

\myheading{Ablation study on different losses.}
 To evaluate the impact of different losses, we train SE-Net with classification loss only, contrastive loss only, and their combination (as in the main experiments) on OSIE and COCO-Search18. Results in \cref{tab:abalation-loss} show that contrastive loss is crucial for capturing subject-specific attention patterns, while classification loss significantly accelerates convergence. On COCO-Search18, combining classification and triplet loss is more effective due to the similarity of attention patterns across subjects. Classification loss alone overlooks these similarities, while triplet loss struggles to find subtle differences. 

\myheading{Ablation study on different modules.}
To ablate the effect of different modules in SE-Net, we train SE-Net under two settings on COCO-Search18. 1) w/o task encoder: we remove task encoder, and direct pass task embedding extracted from text encoder to self-attention in CSE module. 2) w/o duration: do not use duration to update scanpath embedding. Results shown in \cref{tab:abalation-module} indicate the importance of each module in inferring unseen subject embeddings. 
\begin{table}[h!]
    \centering
    \renewcommand{\arraystretch}{1} 
    \setlength{\tabcolsep}{3pt} 
    \resizebox{0.8\columnwidth}{!}{
    \begin{tabular}{p{3cm}p{1.3cm}p{1.3cm}p{1.3cm}}
        \toprule
        Module &  SM $\uparrow$ & MM $\uparrow$ & SED $\downarrow$ \\
        \midrule
        w/o Task Encoder & 0.459 & 0.814 & 2.459 \\
        w/o Duration & 0.446 & 0.814 & 2.539 \\
        ISP-SENet & 0.482 & 0.815 & 2.359 \\
        \bottomrule
    \end{tabular}
    }
    \vskip -0.1in
\caption{Ablation on the effect of different modules of SE-Net on COCO-Search18. \label{tab:abalation-module} } 
\end{table} 
\section{Conclusions and Discussion}
We highlighted the significance of generating personalized scanpath predictions for novel subjects with minimal support samples and introduced the Few-Shot Personalized Scanpath Prediction (FS-PSP) task. To tackle this challenge, we created a pipeline that independently learns subject embeddings to capture individual attention patterns and uses these embeddings to predict scanpaths. SE-Net is engineered to extract subject embeddings, separate individual attention traits from the image content, and facilitate generalization to unseen subjects with very few examples, thereby minimizing biases linked to restricted scenes. By utilizing these robust unseen subject embeddings, our scanpath prediction model markedly outperforms methods that require relearning personalization for new subjects with limited data. Furthermore, we have set a benchmark for FS-PSP, encouraging additional research to develop more comprehensive subject embeddings that are effective across diverse eye-tracking tasks and scenarios. \\
A limitation of SE-Net is, While it distinguishes subjects, it may overemphasize unique traits while overlooking shared patterns beneficial for scanpath prediction. Addressing this could enhance the robustness and generalizability of the method.\\
\textbf{Acknowledgements.} This project is supported by US National Science Foundation grant IIS-2123920.
{
    \small
    \bibliographystyle{ieeenat_fullname}
    \bibliography{main}
}
\clearpage
\setcounter{page}{1}
\maketitlesupplementary

\section{Overview}

This supplementary meterial is arranged as: \\
\begin{itemize}
    \item \cref{sec:supp-implementation} shows the implementation details of ISP-SENet.
    \item \cref{sec:supp-statistics} shows statistics of Tab. 1 in the main paper.
    \item \cref{sec:supp-supp-eval} shows the supplementary evaluation of ISP-SENet. 
    \item \cref{sec:supp-abalation} shows ablation study on more parameters and modules.
    \item \cref{sec:supp-qualitative} shows more qualitative results.\\
In the experiments, unless specified otherwise, we sample the 10-shot support set for 10 times.
\end{itemize}

\section{Implementation Details}
\label{sec:supp-implementation}
\subsection{Feature Extractor $F$}
Humans perceive images through a high-resolution foveal region and a low-resolution peripheral region \cite{curcio1990human}, creating distinct focal and contextual areas. This principle also guides scanpath prediction models \cite{yang2024unifying, yang2022target, li2024imitating}. Following HAT \cite{yang2024unifying}, we encode images and scanpaths using hierarchical feature maps from the image encoder and decoder. The image encoder produces multi-scale feature maps based on ResNet \cite{he2016deep}. To better align with human object-centric attention \cite{scholl2001objects}, deformable attention \cite{zhu2020deformable} pre-trained on segmentation tasks is utilized to generate hierarchical feature maps that capture semantic object information. The output of image decoder is four-scale hierarchical feature maps, where we utilize two feature maps with lowest and highest resolution ($P_l \in \mathbb{R}^{\left(\frac{H}{32} \cdot \frac{W}{32}\right) \times C}$ and $P_h \in \mathbb{R}^{\left(\frac{H}{4} \cdot \frac{W}{4}\right) \times C}$, respectively). $P_l$ is flattened and directly used as image tokens $F_I$, simulating the peripheral region of human attention. We select corresponding location of all fixations from $P_h$ and obtain $F_S \in \mathbb{R}^{L \times C}$, where $L$ is the length of scanpath, resembling the foveated regions of human attention. 

\subsection{Duration}
It should be noted that the duration strategy is exclusively implemented in SE-Net, whereas ISP-SENet utilizes raw durations. \\
This strategy involves categorizing each fixation duration into one of ten bins, ensuring that each bin contains approximately the same number of fixation durations, a method known as quantile-based intervals. This approach is motivated by two main reasons:
\begin{figure}[h]
  \centering
  \includegraphics[width=1\linewidth]{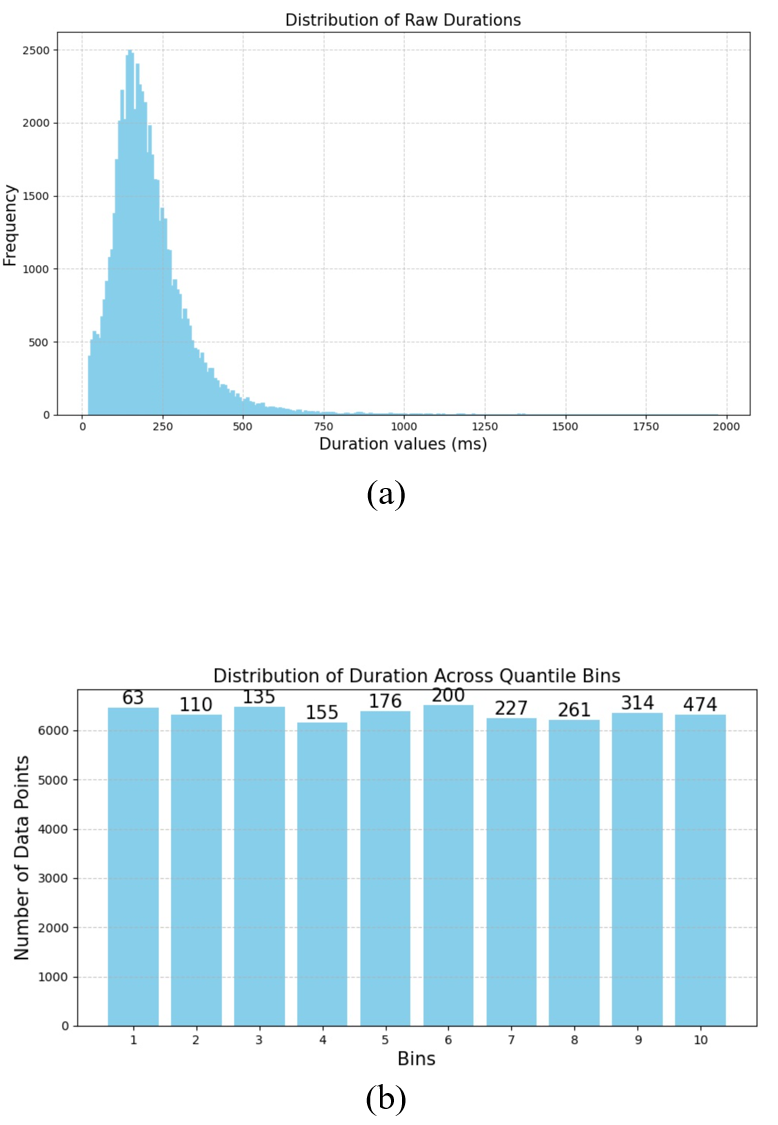}
  \caption{\textbf{Duration distribution.} Figure (a) shows the long-tail distribution of all fixation durations in the base set of seen subjects. Figure (b) visualizes the number of points in each bin. The mean value of each bin is shown on top of each bar. In SE-Net, the bin index replaces the raw duration and is encoded using 1D sinusoidal positional encoding.}
  \label{fig:supp-duration-dist}
\end{figure}
\begin{enumerate}
    \item \textbf{Significance of Duration:} Fixation duration is indicative of the importance attributed to a point in an image, as longer durations generally reflect greater interest by the viewer. By grouping durations into bins, we aim to quantitatively represent the significance, or the underlying importance, of each fixation.
    \item \textbf{Distribution Characteristics:} The fixation durations exhibit a long-tail distribution, as evidenced in \cref{fig:supp-duration-dist} (we collect all fixation durations values across all fixations). Employing a quantile strategy prevents the highly frequent shorter durations from clustering excessively in the initial bins. Instead, it ensures a more balanced distribution across the bins, with larger values being more evenly dispersed among them.
\end{enumerate}
The number of bins is set to 10, and the visualization of each bin's statistics is shown in \cref{fig:supp-duration-dist}. The ablation of duration strategy is discussed in \cref{sec:supp-abalation} and \cref{tab:supp-ablation-duration}.

\section{Statistics of Main Results}
\label{sec:supp-statistics}
\subsection{Margin of Error}
To show the stabilization of our method, we show the margin of error at 95\% confidence level in \cref{tab:supp-std-split1}. It is obtained by sampling the support set 10 times, and ensuring each sampling set is exclusive. From the results, ISP-SENet experienced more stable performance across different support set sampling, while the performance of baselines experienced more variance, suffering from the different image content in the support set. 

\begin{table*}[t]
\centering
\begin{subtable}{1\textwidth}
\centering
\subcaption{OSIE}
\renewcommand{\arraystretch}{0.8} 
\begin{tabular}{llccc}
    \toprule
    $n$-shot & Method & SM $\uparrow$ & MM $\uparrow$ & SED $\downarrow$ \\
    \midrule
    n = 1 & ChenLSTM-ISP & 0.282 {\footnotesize $\pm$ 0.009} & 0.763 {\footnotesize $\pm$ 0.006} & 7.832 {\footnotesize $\pm$ 0.181} \\
          & Gazeformer-ISP & 0.327 {\footnotesize $\pm$ 0.007} & 0.792 {\footnotesize $\pm$ 0.003} & 7.873 {\footnotesize $\pm$ 0.134} \\
          & ChenLSTM-ISP-S & 0.328 {\footnotesize $\pm$ 0.001} & 0.793 {\footnotesize $\pm$ 0.001} & 7.601 {\footnotesize $\pm$ 0.039} \\
          & Gazeformer-ISP-S & 0.354 {\footnotesize $\pm$ 0.000} & 0.801 {\footnotesize $\pm$ 0.000} & 7.503 {\footnotesize $\pm$ 0.003} \\
          & \textbf{ISP-SENet} & \textbf{0.368 {\footnotesize $\pm$ 0.003}} & \textbf{0.805 {\footnotesize $\pm$ 0.002}} & \textbf{7.413 {\footnotesize $\pm$ 0.033}} \\
    \midrule
    n = 5 & ChenLSTM-ISP & 0.319 {\footnotesize $\pm$ 0.005} & 0.773 {\footnotesize $\pm$ 0.004} & 7.855 {\footnotesize $\pm$ 0.116} \\
          & Gazeformer-ISP & 0.340 {\footnotesize $\pm$ 0.003} & 0.791 {\footnotesize $\pm$ 0.002} & 7.920 {\footnotesize $\pm$ 0.082} \\
          & ChenLSTM-ISP-S & 0.329 {\footnotesize $\pm$ 0.001} & 0.801 {\footnotesize $\pm$ 0.000} & 7.499 {\footnotesize $\pm$ 0.028} \\
          & Gazeformer-ISP-S & 0.354 {\footnotesize $\pm$ 0.000} & 0.791 {\footnotesize $\pm$ 0.001} & 7.699 {\footnotesize $\pm$ 0.003} \\
          & \textbf{ISP-SENet} & \textbf{0.376 {\footnotesize $\pm$ 0.002}} & \textbf{0.803 {\footnotesize $\pm$ 0.001}} & \textbf{7.649 {\footnotesize $\pm$ 0.028}} \\
    \midrule
    n = 10 & ChenLSTM-ISP & 0.322 {\footnotesize $\pm$ 0.005} & 0.777 {\footnotesize $\pm$ 0.002} & 7.740 {\footnotesize $\pm$ 0.079} \\
           & Gazeformer-ISP & 0.345 {\footnotesize $\pm$ 0.003} & 0.794 {\footnotesize $\pm$ 0.002} & 7.916 {\footnotesize $\pm$ 0.054} \\
           & ChenLSTM-ISP-S & 0.328 {\footnotesize $\pm$ 0.005} & 0.791 {\footnotesize $\pm$ 0.001} & 7.637 {\footnotesize $\pm$ 0.060} \\
           & Gazeformer-ISP-S & 0.354 {\footnotesize $\pm$ 0.000} & 0.802 {\footnotesize $\pm$ 0.000} & 7.505 {\footnotesize $\pm$ 0.003} \\
           & \textbf{ISP-SENet} & \textbf{0.375 {\footnotesize $\pm$ 0.001}} & \textbf{0.803 {\footnotesize $\pm$ 0.001}} & \textbf{7.318 {\footnotesize $\pm$ 0.017}} \\
    \bottomrule
\end{tabular}
\label{tab:supp-osie-split1}
\end{subtable}

\renewcommand{\arraystretch}{0.8} 
\begin{subtable}{1\textwidth}
\centering
\subcaption{COCO-FreeView}
\begin{tabular}{llccc}
    \toprule
    $n$-shot & Method & SM $\uparrow$ & MM $\uparrow$ & SED $\downarrow$ \\
    \midrule
    n = 1 & ChenLSTM-ISP & 0.287 {\footnotesize $\pm$ 0.014} & 0.805 {\footnotesize $\pm$ 0.003} & 13.307 {\footnotesize $\pm$ 0.195} \\
          & Gazeformer-ISP & 0.244 {\footnotesize $\pm$ 0.021} & 0.787 {\footnotesize $\pm$ 0.011} & 15.118 {\footnotesize $\pm$ 0.510} \\
          & ChenLSTM-ISP-S & 0.339 {\footnotesize $\pm$ 0.000} & 0.814 {\footnotesize $\pm$ 0.000} & 12.523 {\footnotesize $\pm$ 0.029} \\
           & Gazeformer-ISP-S & 0.333 {\footnotesize $\pm$ 0.000} & 0.817 {\footnotesize $\pm$ 0.000} & 12.538 {\footnotesize $\pm$ 0.012} \\
          & \textbf{ISP-SENet} & \textbf{0.369 {\footnotesize $\pm$ 0.002}} & \textbf{0.832 {\footnotesize $\pm$ 0.001}} & \textbf{12.227 {\footnotesize $\pm$ 0.134}} \\
    \midrule
    n = 5 & ChenLSTM-ISP & 0.320 {\footnotesize $\pm$ 0.009} & 0.815 {\footnotesize $\pm$ 0.005} & 12.950 {\footnotesize $\pm$ 0.190} \\
          & Gazeformer-ISP & 0.286 {\footnotesize $\pm$ 0.012} & 0.800 {\footnotesize $\pm$ 0.005} & 14.630 {\footnotesize $\pm$ 0.310} \\
          & ChenLSTM-ISP-S & 0.338 {\footnotesize $\pm$ 0.000} & 0.814 {\footnotesize $\pm$ 0.000} & 12.540 {\footnotesize $\pm$ 0.023} \\
          & Gazeformer-ISP-S & 0.333 {\footnotesize $\pm$ 0.000} & 0.817 {\footnotesize $\pm$ 0.000} & 12.539 {\footnotesize $\pm$ 0.008} \\
          & \textbf{ISP-SENet} & \textbf{0.368 {\footnotesize $\pm$ 0.001}} & \textbf{0.829 {\footnotesize $\pm$ 0.001}} & \textbf{12.017 {\footnotesize $\pm$ 0.058}} \\
    \midrule
    n = 10 & ChenLSTM-ISP & 0.323 {\footnotesize $\pm$ 0.010} & 0.819 {\footnotesize $\pm$ 0.005} & 12.541 {\footnotesize $\pm$ 0.114} \\
           & Gazeformer-ISP & 0.317 {\footnotesize $\pm$ 0.002} & 0.805 {\footnotesize $\pm$ 0.002} & 14.224 {\footnotesize $\pm$ 0.207} \\
           & ChenLSTM-ISP-S & 0.340 {\footnotesize $\pm$ 0.000} & 0.814 {\footnotesize $\pm$ 0.000} & 12.532 {\footnotesize $\pm$ 0.025} \\
           & Gazeformer-ISP-S & 0.333 {\footnotesize $\pm$ 0.000} & 0.816 {\footnotesize $\pm$ 0.000} & 12.545 {\footnotesize $\pm$ 0.006} \\
           & \textbf{ISP-SENet} & \textbf{0.367 {\footnotesize $\pm$ 0.001}} & \textbf{0.828 {\footnotesize $\pm$ 0.001}} & \textbf{11.956 {\footnotesize $\pm$ 0.010}}  \\
    \bottomrule
\end{tabular}
\label{tab:supp-coco-freeview-split1}
\end{subtable}

\renewcommand{\arraystretch}{0.8} 
\begin{subtable}{1\textwidth}
\centering
\subcaption{COCO-Search18}
\begin{tabular}{llccc}
    \toprule
    $n$-shot & Method & SM $\uparrow$ & MM $\uparrow$ & SED $\downarrow$ \\
    \midrule
    n = 1 & ChenLSTM-ISP & 0.371 {\footnotesize $\pm$ 0.024} & 0.760 {\footnotesize $\pm$ 0.029} & 2.756 {\footnotesize $\pm$ 0.464}  \\
          & Gazeformer-ISP & 0.342 {\footnotesize $\pm$ 0.018} & 0.770 {\footnotesize $\pm$ 0.008} & 2.818 {\footnotesize $\pm$ 0.216}  \\
          & ChenLSTM-ISP-S & 0.448 {\footnotesize $\pm$ 0.000} & 0.803 {\footnotesize $\pm$ 0.001} & 2.394 {\footnotesize $\pm$ 0.013}  \\
          & Gazeformer-ISP-S & 0.446 {\footnotesize $\pm$ 0.001} & 0.802 {\footnotesize $\pm$ 0.001} & 2.463 {\footnotesize $\pm$ 0.002}  \\
          & \textbf{ISP-SENet} & \textbf{0.475 {\footnotesize $\pm$ 0.007}} & \textbf{0.814 {\footnotesize $\pm$ 0.001}} & \textbf{2.333 {\footnotesize $\pm$ 0.063}} \\
    \midrule
    n = 5 & ChenLSTM-ISP & 0.386 {\footnotesize $\pm$ 0.015} & 0.773 {\footnotesize $\pm$ 0.008} & 2.489 {\footnotesize $\pm$ 0.058}  \\
          & Gazeformer-ISP & 0.353 {\footnotesize $\pm$ 0.028} & 0.774 {\footnotesize $\pm$ 0.011} & 2.980 {\footnotesize $\pm$ 0.292}  \\
          & ChenLSTM-ISP-S & 0.449 {\footnotesize $\pm$ 0.001} & 0.803 {\footnotesize $\pm$ 0.001} & 2.380 {\footnotesize $\pm$ 0.014}  \\
          & Gazeformer-ISP-S & 0.445 {\footnotesize $\pm$ 0.001} & 0.803 {\footnotesize $\pm$ 0.001} & 2.457 {\footnotesize $\pm$ 0.002}  \\
          & \textbf{ISP-SENet} & \textbf{0.484 {\footnotesize $\pm$ 0.005}} & \textbf{0.815 {\footnotesize $\pm$ 0.001}} & \textbf{2.354 {\footnotesize $\pm$ 0.044}} \\
    \midrule
    n = 10 & ChenLSTM-ISP & 0.393 {\footnotesize $\pm$ 0.006} & 0.781 {\footnotesize $\pm$ 0.004} & 2.394 {\footnotesize $\pm$ 0.038}  \\
           & Gazeformer-ISP & 0.370 {\footnotesize $\pm$ 0.007} & 0.785 {\footnotesize $\pm$ 0.006} & 2.765 {\footnotesize $\pm$ 0.128}  \\
           & ChenLSTM-ISP-S & 0.449 {\footnotesize $\pm$ 0.000} & 0.803 {\footnotesize $\pm$ 0.001} & 2.379 {\footnotesize $\pm$ 0.019}  \\
           & Gazeformer-ISP-S & 0.446 {\footnotesize $\pm$ 0.001} & 0.802 {\footnotesize $\pm$ 0.001} & 2.464 {\footnotesize $\pm$ 0.002}  \\
           & \textbf{ISP-SENet} & \textbf{0.482 {\footnotesize $\pm$ 0.002}} & \textbf{0.815 {\footnotesize $\pm$ 0.001}} & \textbf{2.359 {\footnotesize $\pm$ 0.019}} \\
    \bottomrule
\end{tabular}
\label{tab:supp-coco-search18-split1}
\end{subtable}
\vskip -0.1in
\caption{Margin of error for Tab. 1 in the main paper.\label{tab:supp-std-split1}}

\end{table*}
\subsection{Second Seen-Unseen Split}
To ensure the result is not biased on subjects due to the model's ability may various on different subjects, we conduct one more split of seen-unseen subjects, which still follows the rule of 70\% seen and 30\% unseen. To specify, 10 seen subjects and 5 unseen subjects for OSIE, 7 seen and 3 unseen subjects for both COCO-Search18 and COCO-FreeView. We ensure this second split contains different unseen subjects compared with the split shown in the main paper. The result is shown in \cref{tab:supp-std-split2}. From the results in \cref{tab:supp-std-split1} and \cref{tab:supp-std-split2}, we observe that ISP-SENet demonstrates stability across different seen-unseen splits, as indicated by relatively consistent performance metrics. In contrast, the performance variations in the two baselines are significant. This difference is largely attributed to the fine-tuning process, where performance is heavily dependent on the small support set. With only 10 images from each subject, substantial variation arises due to biases in image content and the specific human attention related to individual scenes. These observations further underscore the limitations of existing methods in few-shot settings.

\begin{table*}[ht]
\centering
\begin{subtable}{1\textwidth}
\centering
\subcaption{OSIE}
\begin{tabular}{lccc}
    \toprule
    Method & SM $\uparrow$ & MM $\uparrow$ & SED $\downarrow$ \\
    \midrule
    ChenLSTM-ISP & 0.288 {\footnotesize $\pm$ 0.009} & 0.780 {\footnotesize $\pm$ 0.004} & \textbf{7.350 {\footnotesize $\pm$ 0.127}} \\
    Gazeformer-ISP & 0.318 {\footnotesize $\pm$ 0.006} & 0.789 {\footnotesize $\pm$ 0.002} & 8.363 {\footnotesize $\pm$ 0.155} \\
    \textbf{ISP-SENet} & \textbf{0.384 {\footnotesize $\pm$ 0.001}} & \textbf{0.813 {\footnotesize $\pm$ 0.001}} & 7.460 {\footnotesize $\pm$ 0.022} \\
    \bottomrule
\end{tabular}
\label{tab:supp-osie-split2}
\end{subtable}

\vspace{1em} 

\begin{subtable}{1\textwidth}
\centering
\subcaption{COCO-FreeView}
\begin{tabular}{lccc}
    \toprule
    Method & SM $\uparrow$ & MM $\uparrow$ & SED $\downarrow$ \\
    \midrule
    ChenLSTM-ISP & 0.296 {\footnotesize $\pm$ 0.007} & 0.823 {\footnotesize $\pm$ 0.001} & 12.534 {\footnotesize $\pm$ 0.030} \\
    Gazeformer-ISP & 0.275 {\footnotesize $\pm$ 0.008} & 0.801 {\footnotesize $\pm$ 0.006} & 14.266 {\footnotesize $\pm$ 0.286} \\
    \textbf{ISP-SENet} & \textbf{0.364 {\footnotesize $\pm$ 0.001}} & \textbf{0.835 {\footnotesize $\pm$ 0.001}} & \textbf{12.342 {\footnotesize $\pm$ 0.019}}  \\
    \bottomrule
\end{tabular}
\label{tab:supp-coco-freeview-split2}
\end{subtable}

\vspace{1em} 

\begin{subtable}{1\textwidth}
\centering
\subcaption{COCO-Search18}
\begin{tabular}{lccc}
    \toprule
     Method & SM $\uparrow$ & MM $\uparrow$ & SED $\downarrow$ \\
    \midrule
    ChenLSTM-ISP & 0.333 {\footnotesize $\pm$ 0.006} & 0.766 {\footnotesize $\pm$ 0.006} & 2.712 {\footnotesize $\pm$ 0.042}  \\
    Gazeformer-ISP & 0.403 {\footnotesize $\pm$ 0.010} & 0.803 {\footnotesize $\pm$ 0.005} & 2.734 {\footnotesize $\pm$ 0.116}  \\
    \textbf{ISP-SENet} & \textbf{0.465 {\footnotesize $\pm$ 0.001}} & \textbf{0.812 {\footnotesize $\pm$ 0.001}} & \textbf{2.286 {\footnotesize $\pm$ 0.020}} \\
    \bottomrule
\end{tabular}
\label{tab:supp-coco-search18-split2}
\end{subtable}
\vskip -0.1in
\caption{Results from the second seen-unseen split under the 10-shot setting. The unseen subject set in this split is distinct from the unseen set used in the main paper's split.\label{tab:supp-std-split2}}
\end{table*}

\section{Supplementary Evaluation}
\label{sec:supp-supp-eval}
In this section, we evaluate the performance of ISP-SENet in three aspect: \\
In \cref{sec:supp-seen}, we compare the performance of ISP-SENet and baselines on seen subject.\\
In \cref{sec:supp-diff-scanpath-model}, we use subject embeddings learned from SE-Net to replace the subject embedding of ChenLSTM-ISP, and compare with baseline fine-tuned on full training set of unseen subjects.\\
In \cref{sec:supp-dintinct}, we develop a new evaluation method to validate that ISP-SENet can distinct different subject embeddings.

\subsection{Results on Seen Set}
\label{sec:supp-seen}
In \cref{tab:supp-seen}, we evaluate the performance of ISP-SENet and baselines on seen subjects, same as the split defined in the main paper. Notably, although the subject embeddings generated by SE-Net are frozen during the training process of ISP-SENet, indicating that they are not tailored for scanpath prediction, the performance on COCO-Search18 is significantly better. Moreover, it achieved comparable results with OSIE and COCO-FreeView. This suggests that the seen subject embeddings learned by SE-Net, despite being optimized for distinguishing different subjects rather than specifically for scanpath prediction, effectively retain individual attention traits and excel in personalized scanpath prediction.
\begin{table*}[h!]
    \centering
    \renewcommand{\arraystretch}{1} 
    \setlength{\tabcolsep}{4pt} 
    
    \begin{tabular}{lccccccccc}
        \toprule
        \multirow{2}{*}{Methods} & \multicolumn{3}{c}{OSIE} & \multicolumn{3}{c}{COCO-FreeView} & \multicolumn{3}{c}{COCO-Search18}\\
        \cmidrule(lr){2-4} \cmidrule(lr){5-7} \cmidrule(lr){8-10}
         & SM $\uparrow$ & MM $\uparrow$ & SED $\downarrow$ & SM $\uparrow$ & MM $\uparrow$ & SED $\downarrow$ & SM $\uparrow$ & MM $\uparrow$ & SED $\downarrow$ \\
        \midrule
        ChenLSTM-ISP& 0.373 & 0.814 & 7.171 & 0.373 & 0.828 & 12.126 & 0.475 & 0.820 & 2.128 \\
        Gazeformer-ISP & 0.382 & 0.813 & \textbf{7.077} & \textbf{0.380} & \textbf{0.835} & \textbf{11.707} & 0.480 & 0.815 & 2.204 \\
        \textbf{ISP-SENet} & \textbf{0.382} & \textbf{0.816} & 7.127 & 0.375 & 0.833 & 11.872 & \textbf{0.517} & \textbf{0.825} & \textbf{2.086} \\
        \bottomrule
    \end{tabular}
    \vskip -0.1in
\caption{Performance Comparison of methods on seen subjects. All methods are trained on all training data of seen subjects, and test on the test set of seen subjects. \label{tab:supp-seen} }         
\end{table*}

\subsection{ISP-SENet with Different Scanpath Prediction Models}
\label{sec:supp-diff-scanpath-model}
To demonstrate the adaptability of the subject embeddings learned from SE-Net across different scanpath prediction models, we substituted the original subject embeddings in ChenLSTM-ISP with those learned from SE-Net. The performance of this configuration, referred to as ISP-SENet(ChenLSTM-ISP), is shown in the fourth row of \cref{tab:supp-chenlstm-senet}.\\
Further, to compare the performance of ISP-SENet with baselines, we fine-tuned both Gazeformer-ISP and ChenLSTM-ISP on the complete training set for unseen subjects. The results, labeled as Gazeformer-ISP-FT and ChenLSTM-ISP-FT, are presented in the first and third rows of \cref{tab:supp-chenlstm-senet}.\\
These results highlight that, without any fine-tuning on unseen subjects, ISP-SENet achieves comparable results compared with baseline models, which are fully fine-tuned on full training set of unseen subjects. 
\begin{table*}[h!]
    \centering
    \renewcommand{\arraystretch}{1} 
    \setlength{\tabcolsep}{4pt} 
    
    \begin{tabular}{lccccccccc}
        \toprule
        \multirow{2}{*}{Methods} & \multicolumn{3}{c}{OSIE} & \multicolumn{3}{c}{COCO-FreeView} & \multicolumn{3}{c}{COCO-Search18}\\
        \cmidrule(lr){2-4} \cmidrule(lr){5-7} \cmidrule(lr){8-10}
         & SM $\uparrow$ & MM $\uparrow$ & SED $\downarrow$ & SM $\uparrow$ & MM $\uparrow$ & SED $\downarrow$ & SM $\uparrow$ & MM $\uparrow$ & SED $\downarrow$ \\
        \midrule
        Gazeformer-ISP-FT & 0.372 & 0.803 & 7.614 & 0.383 & 0.834 & 11.443 & 0.479 & 0.815 & 2.330 \\
        ISP-SENet (Gazeformer-ISP) & 0.375 & 0.803 & 7.318 & 0.367 & 0.828 & 11.956 & 0.482 & 0.815 & 2.359 \\
        \midrule
        ChenLSTM-ISP-FT & 0.371 & 0.801 & 7.449 & 0.387 & 0.832 & 11.422 & 0.475 & 0.813 & 2.159 \\
        ISP-SENet (ChenLSTM-ISP) & 0.369 & 0.800 & 7.574 & 0.366 & 0.824 & 12.241 & 0.467 & 0.810 & 2.272 \\
        \bottomrule
    \end{tabular}
    \vskip -0.1in
\caption{ISP-SENet with Different Scanpath Prediction Models, and comparsion between ISP-SENet without fine-tuning on unseen subjects, with ISP\cite{chen2024beyond} fine-tuned on full training set of unseen subjects. \label{tab:supp-chenlstm-senet} }         
\end{table*}

\subsection{Cross-subject embedding Evaluation}
\label{sec:supp-dintinct}
To confirm that our unseen subject embeddings capture unique attention patterns rather than a global optimum applicable to all subjects, we implement a cross-subject embedding evaluation. For each unseen subject $u_k$, we first calculate the SM, MM, and SED metrics using the subject's own embedding $e_k$ for predicting scanpaths on the query set. Then we replace $e_k$ with embeddings $e_i$ from $m$ different subjects $u_i$, where $u_i \in U_{\text{unseen}, i \neq k}$, and compute the average SM, MM, and SED. The differences in these metrics underscore the uniqueness of each embedding. For simplicity, we define $m=3$ and randomly sampled 5 different support sets.

In \cref{tab:supp-distinct}, the symbol $\times$ represents that the subject embedding and prediction correspond to the same subject, while $\checkmark$ indicates they belong to different subjects. To better understand the model performance of cross-subject embedding evaluation, we include comparisons with ISP(Seen) and ISP-SENet(Seen). \textbf{ISP(Seen)} evaluates Gazeformer-ISP's ability to differentiate among embeddings of seen subjects. As ISP-SENet is built upon Gazeformer-ISP, the distinction ability of these two models will not have significant differences. \textbf{ISP-SENet(Seen)} assesses ISP-SENet's cross-subject embedding performance on seen subjects, indicative of the potential upper limit of our model's discriminative capability. \textbf{ISP-SENet(Unseen)} represents our cross-subject embedding evaluation for unseen subjects. The results demonstrate that ISP-SENet's capacity to distinguish unseen subjects exceeds the baseline's performance with seen subjects and is comparable with ISP-SENet's performance on seen subjects.

\begin{table}[h!]
    \centering
    \renewcommand{\arraystretch}{1.2} 
    \setlength{\tabcolsep}{3pt} 
     \resizebox{0.95\columnwidth}{!}{
    \begin{tabular}{c|c|ccc}
        \multirow{2}{*}{Method} & \multirow{2}{*}{cross-subject embedding} & \multicolumn{3}{c}{\textbf{OSIE}}\\
        \cmidrule(lr){3-5}
        & & SM $\uparrow$ & MM $\uparrow$ & SED $\downarrow$\\
        \midrule

        \multirow{2}{*}{ISP(Seen)}
        & $\times$ & 0.386 & 0.814 & 7.003\\
        & $\checkmark$ & 0.379 & 0.812 & 7.163\\
        \midrule
        \multirow{2}{*}{ISP-SENet(Seen)} 
        & $\times$ & 0.387 & 0.815 & 7.009\\
        & $\checkmark$ & 0.373 & 0.810 & 7.360\\
        \midrule
        \multirow{2}{*}{ISP-SENet(Unseen)} 
        & $\times$ & 0.376 & 0.802 & 7.286\\
        & $\checkmark$ & 0.361 & 0.800 & 7.340\\
        \midrule
    \end{tabular}
    }
    \vskip -0.1in
\caption{Cross-embedding evaluation on the distinction ability between subject embeddings. The symbol $\times$ denotes that the subject embedding and prediction correspond to the same subject, while $\checkmark$ indicates that the subject embeddings and prediction belong to different subjects.\label{tab:supp-distinct}}     
\end{table}

\subsection{Quantatitive results on visual-task encoder}
To demonstrate that the visual-task encoder effectively captures the alignment between the task and image content, we evaluate the similarity between its cross-attention maps and the ground truth bounding boxes using Correlation Coefficient (CC) and AUC. For SE-Net, we achieve a CC of 0.31 and an AUC of 0.76. While the CC is sensitive to false positives—such as attention allocated to relevant peripheral objects—our model still outperforms ChenLSTM-ISP’s task-guidance map $m_0$(CC = 0.07, AUC = 0.63), averaged across channels. This indicates stronger target understanding, despite our model not being explicitly designed for object detection.

\subsection{More analysis on size of base set}
In \cref{tab:supp-num-seen-subject}, we analyze the impact of varying the number of seen subjects in the base set during training on COCO-Search18, supplementing the results in Table 4 of the main paper. We consistently select one subject as unseen and vary the number of seen subjects selected from the remaining ones. With fewer seen subjects, SE-Net struggles to infer the attention traits of new subjects based on its learned experience. Performance improves when increasing the number of seen subjects from 7 (as in the main paper) to 9, suggesting that ISP-SE-Net benefits from additional subjects.
\begin{table}[h!]
    \centering
    \setlength{\tabcolsep}{2pt}  
    \resizebox{0.8\columnwidth}{!}{
    \begin{tabular}{p{4.0cm}p{1.0cm}p{1.0cm}p{1.0cm}}
        \toprule
        num seen &  SM $\uparrow$ & MM $\uparrow$ & SED $\uparrow$ \\
        \midrule
        4(40\%) & 0.472 & 0.819 & 2.542 \\
        9(90\%) & \textbf{0.489} & \textbf{0.826} & \textbf{2.145} \\
        Ours(70\%) & 0.487 & 0.823 & 2.333 \\
        \bottomrule
    \end{tabular}}
    \vskip -0.1in
\caption{ Performance comparison of 10-shot setting on COCO-Search18 with different numbers of seen subjects in training stage. \label{tab:supp-num-seen-subject}} 
\end{table}

\section{More Ablation Results}
\label{sec:supp-abalation}

\subsection{Duration}
In \cref{tab:supp-ablation-duration}, we ablate the effect of duration encoding strategy on OSIE in three settings: (1) No duration encoding in scanpath embeddings. (2) Encoding raw duration without assigning bin index. (3) Assigning durations to 10 bins of equal width, without employing the quantile strategy. (4) Assigning durations to 100 bins using the quantile strategy. (5) Assigning durations to 300 bins using the quantile strategy. (6) Assigning durations to 10 bins using the quantile strategy as in the main paper. \\
The result indicates that: (1) Duration is crucial for understanding subject attention traits. (2) Raw durations offer limited information as they introduce redundancy. For instance, 200 ms and 201 ms are treated as distinct durations, despite their negligible difference, which does not accurately reflect varying importance levels between two fixations. (3) Without the quantile strategy, the bins fail to manage the long-tail effect effectively, resulting in sparse distribution of higher durations across most bins while smaller durations crowd into a few bins due to their higher frequency. (4) and (5) demonstrate how varying the number of bins impacts performance. 
\begin{table}[h!]
    \centering
    \renewcommand{\arraystretch}{1} 
    \setlength{\tabcolsep}{3pt} 
    \begin{tabular}{p{3cm}p{1.3cm}p{1.3cm}p{1.3cm}}
        \toprule
        Duration Strategy &  SM $\uparrow$ & MM $\uparrow$ & SED $\downarrow$ \\
        \midrule
        w/o Duration & 0.365 & 0.797 & 7.634 \\
        Raw duration & 0.367 & 0.800 & 7.534 \\
        Uniform bin width & 0.369 & 0.799 & 7.431 \\
        100 bins & 0.374 & 0.801 & 7.377 \\
        300 bins & 0.370 & 0.801 & 7.474 \\
        \textbf{ISP-SENet} (10 bins) & \textbf{0.375} & \textbf{0.803} & \textbf{7.318} \\
        \bottomrule
    \end{tabular}
    \vskip -0.1in
\caption{Ablation on performance with different duration encoding strategy on OSIE. \label{tab:supp-ablation-duration}}
\end{table} 

\subsection{Margin in Contrastive Loss}
We ablate the effect of different margins $m$ in the contrastive loss. The margin is a predefined threshold that specifies how much farther the negative example should be from the anchor compared to the positive example. As shown in \cref{tab:supp-ablation-margin}, lower or higher margins decrease the prediction performance. A possible reason is lower margin prevents SE-Net from distinguishing different subjects. \\
The performance decreases associated with a higher margin can be attributed to the characteristics of human attention. Despite differences among subjects, their scanpaths often share similarities, such as a focus on foreground objects like humans. Such similarity is critical for SE-Net to learn the subject embedding, and plays a key role in infering embeddings for unseen subjects. We anticipate that embeddings for unseen subjects will benefit from seen subjects with similar attention patterns. Thus, setting a higher margin may overlook these essential similarities. \\
\cref{tab:supp-ablation-margin} shows that, for COCO-Search18 dataset where viewing patterns between people are more similar (higher Human Consistency(HC)\cite{chen2021predicting}), a smaller margin of 1 performs better. For OSIE dataset with more diverse patterns (lower HC), a larger margin of 5 performs better. Also the effect of margin is more significant for higher HC. 

\begin{table}[h!]
    \centering
    \renewcommand{\arraystretch}{1} 
    \setlength{\tabcolsep}{4pt} 
    
    \resizebox{0.9\columnwidth}{!}{
    \begin{tabular}{lcccccc}
        \toprule
        \multirow{2}{*}{margin} & \multicolumn{3}{c}{OSIE(HC=0.39)} & \multicolumn{3}{c}{COCO-Search18(HC=0.52)} \\
        \cmidrule(lr){2-4} \cmidrule(lr){5-7}
         & SM $\uparrow$ & MM $\uparrow$ & SED $\downarrow$ & SM $\uparrow$ & MM $\uparrow$ & SED $\downarrow$ \\
        \midrule
        1 & 0.369 & 0.804 & 7.506 & \textbf{0.482} & \textbf{0.815} & \textbf{2.359} \\
        5 & \textbf{0.375} & 0.803 & \textbf{7.318} & 0.467 & 0.815 & 2.455 \\
        10 & 0.367 & \textbf{0.809} & 7.546 & 0.445 & 0.813 & 2.563 \\
        \bottomrule
    \end{tabular}
    }
    \vskip -0.1in
\caption{Ablation on performance with different $m$ in contrastive loss on OSIE and COCO-Search18. In the main paper, we use $m=5$ for OSIE and $m=1$ for COCO-Search18. \label{tab:supp-ablation-margin}}     
\end{table}


\subsection{Embedding Dimension}
In \cref{tab:supp-embed-dim} we explore different embedding dimensions of SE-Net and ISP-SENet. All layers in SE-Net and ISP-SENet shares the same embedding dimensions. The results indicate that varying the embedding dimensions does not significantly impact performance.
\begin{table}[h!]
    \centering
    \renewcommand{\arraystretch}{1} 
    \setlength{\tabcolsep}{3pt} 
    \resizebox{0.95\columnwidth}{!}{
    \begin{tabular}{C{4cm}p{1.3cm}p{1.3cm}p{1.3cm}}
        \toprule
        Embedding Dimension &  SM $\uparrow$ & MM $\uparrow$ & SED $\downarrow$ \\
        \midrule
       128 & 0.374 & \textbf{0.804} & 7.324 \\
       384 & \textbf{0.375} & 0.803 & \textbf{7.318} \\
        \bottomrule
    \end{tabular}
    }
    \vskip -0.1in
\caption{Ablation on performance with different embedding dimensions. \label{tab:supp-embed-dim}}
\end{table}
\section{Qualitative Results}
\label{sec:supp-qualitative}
\begin{figure*}[t]
  \centering
  \includegraphics[width=1.0\linewidth]{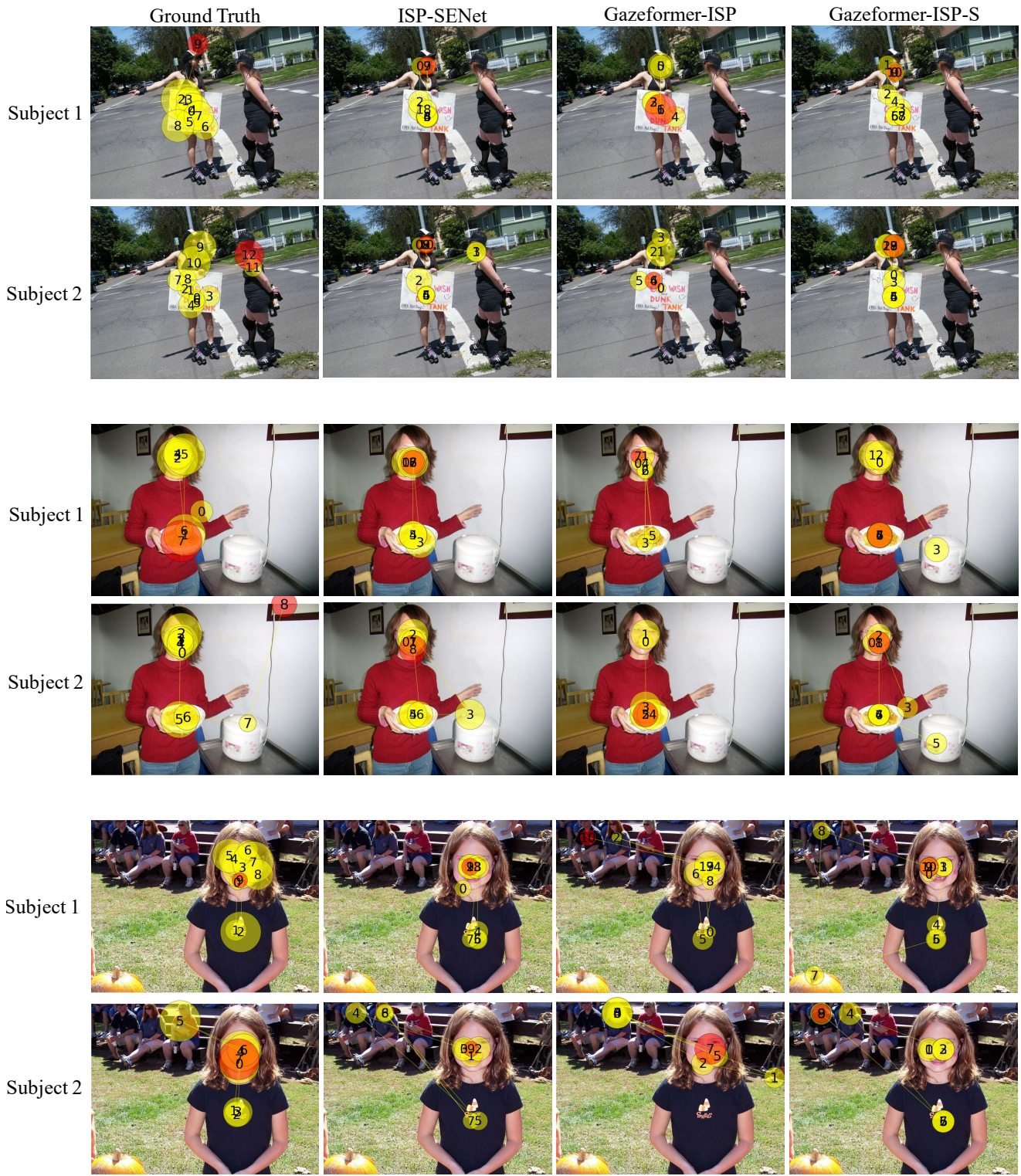}
  \caption{\textbf{More Qualitative examples of scanpath prediction for different unseen subjects on OSIE. } GT is the ground truth scanpaths of different unseen subjects. \textcolor{red}{Red} circle is the end fixation. Each two rows of the same image are scanpaths belonging to two different unseen subjects.}
  \label{fig:supp-vis-osie}
\end{figure*}

\begin{figure*}[t]
  \centering
  \includegraphics[width=1.0\linewidth]{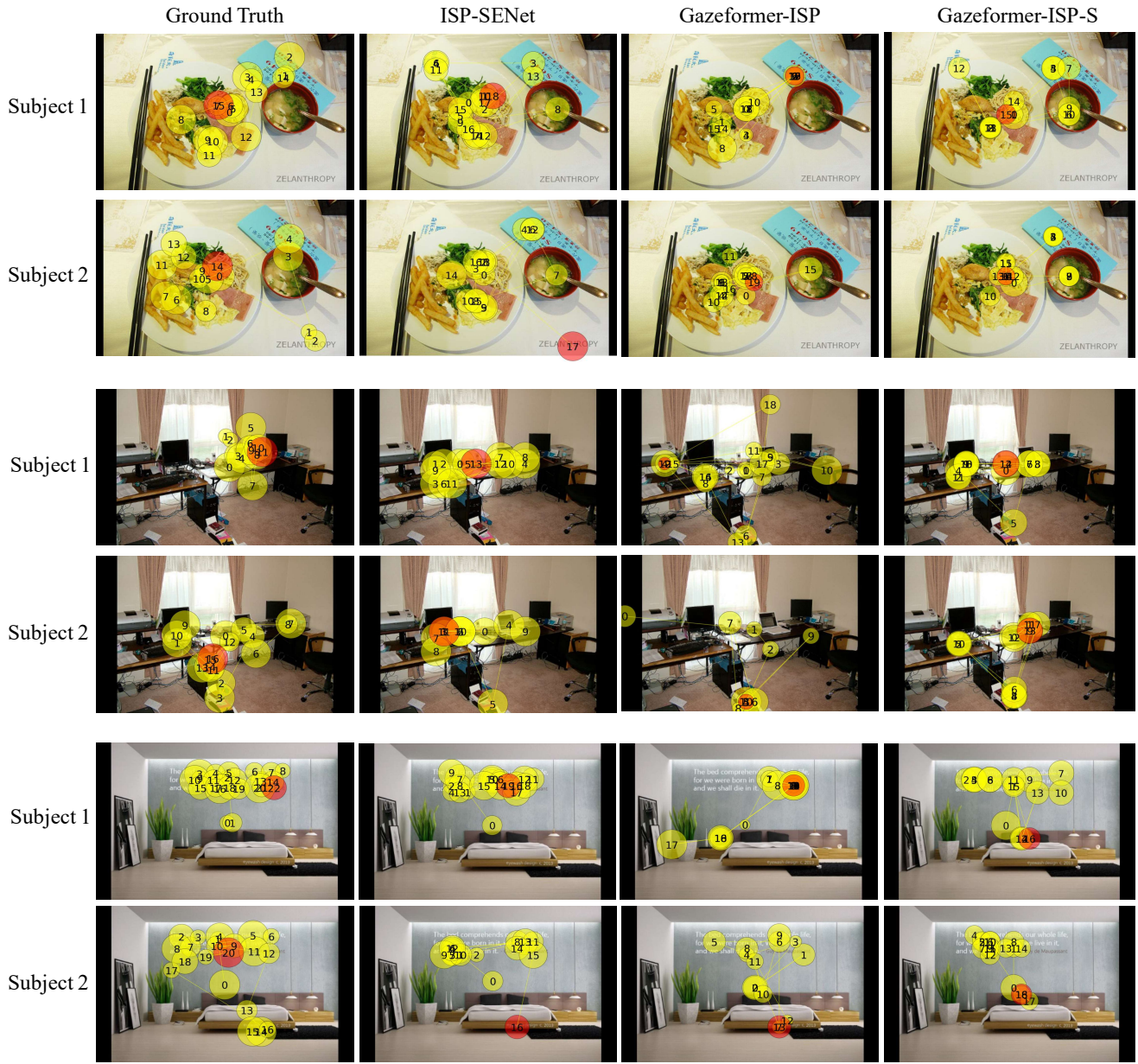}
  \caption{\textbf{More Qualitative examples of scanpath prediction for different unseen subjects on COCO-FreeView. } GT is the ground truth scanpaths of different unseen subjects. \textcolor{red}{Red} circle is the end fixation. Each two rows of the same image are scanpaths belonging to two different unseen subjects.}
  \label{fig:supp-vis-cocofv}
\end{figure*}

\begin{figure*}[h]
  \centering
  \includegraphics[width=1.0\linewidth]{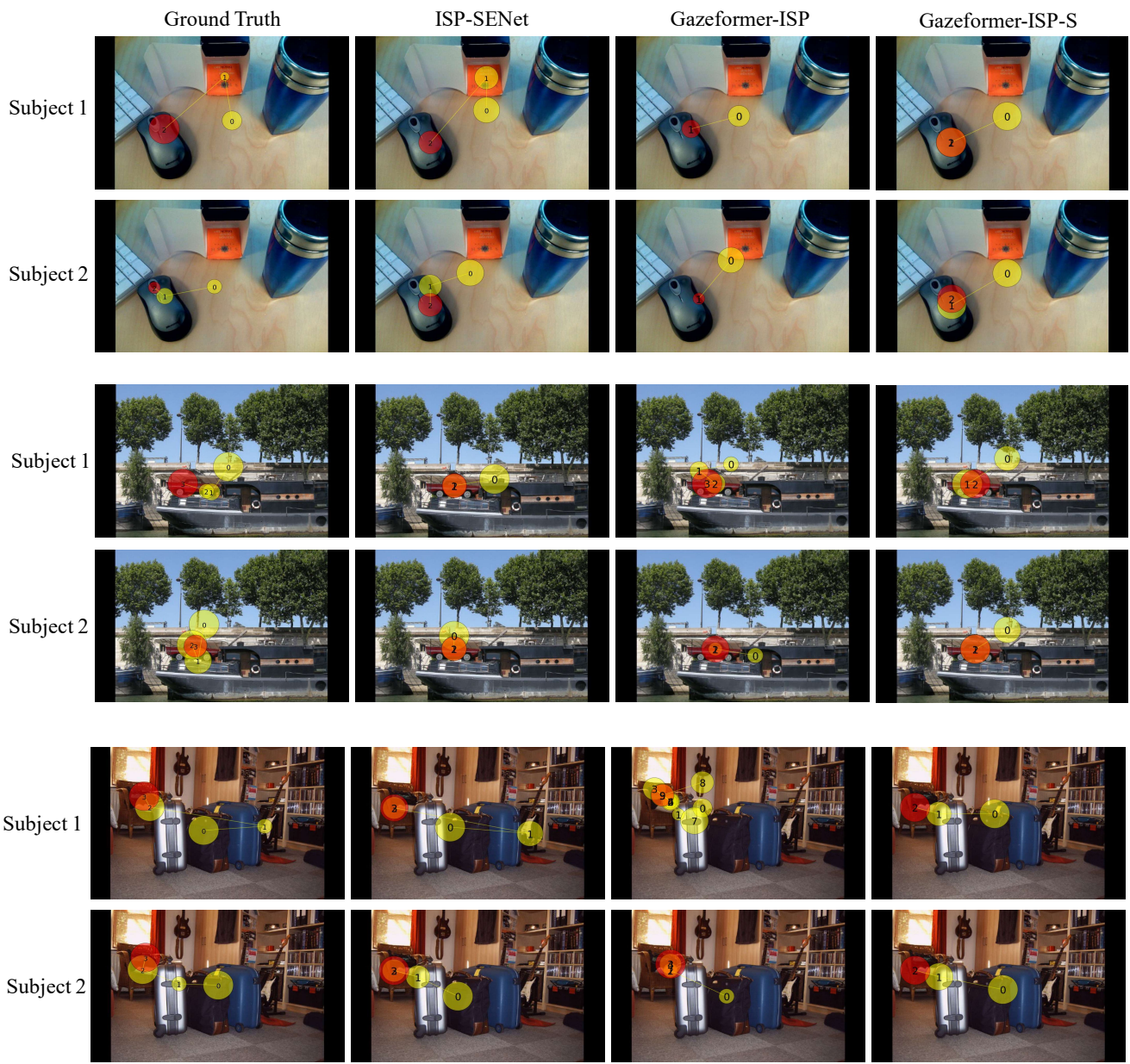}
  \caption{\textbf{More Qualitative examples of scanpath prediction for different unseen subjects on COCO-Search18. } GT is the ground truth scanpaths of different unseen subjects. \textcolor{red}{Red} circle is the end fixation. Each two rows of the same image are scanpaths belonging to two different unseen subjects. The search targets are \textbf{mouse, car, chair}, respectively.}
  \label{fig:supp-vis-coco18}
\end{figure*}

We show more qualitative results of ISP-SENet on OSIE, COCO-FreeView and COCO-Search18 in \cref{fig:supp-vis-osie}, \cref{fig:supp-vis-cocofv}, \cref{fig:supp-vis-coco18}. In most cases, ISP-SE-Net successfully captures the variation in viewed objects across different subjects. Notably, on COCO-FreeView, it learns global attention patterns such as centralized or scattered focus. In the search task, our model also identifies subjects influenced by distractions, outperforming the baselines in capturing such behaviors.


\end{document}